\documentclass[conference,a4paper]{IEEEtran}
\IEEEoverridecommandlockouts
\usepackage{cite}
\usepackage{amsmath,amssymb,amsfonts}
\usepackage{algorithmic}
\usepackage[dvips]{graphicx}
\usepackage{textcomp}
\usepackage{xcolor}

\usepackage{subfigure}
\usepackage{url}
\usepackage{listings}
\usepackage{algorithm}
\usepackage{algorithmic}
\usepackage{multirow}
\usepackage{diagbox}

\newcommand{\bvec}[1]{\mbox{\boldmath $#1$}}

\def\BibTeX{{\rm B\kern-.05em{\sc i\kern-.025em b}\kern-.08em
    T\kern-.1667em\lower.7ex\hbox{E}\kern-.125emX}}
\begin{document}

\title{An Adaptive Structural Learning of Deep Belief Network for Image-based Crack Detection in Concrete Structures Using SDNET2018
\thanks{\copyright 2020 IEEE. Personal use of this material is permitted. Permission from IEEE must be obtained for all other uses, in any current or future media, including reprinting/republishing this material for advertising or promotional purposes, creating new collective works, for resale or redistribution to servers or lists, or reuse of any copyrighted component of this work in other works.}
}
\author{
\IEEEauthorblockN{Shin Kamada}
\IEEEauthorblockA{Advanced Artificial Intelligence Project Research Center,\\
Research Organization of Regional Oriented Studies,\\
Prefectural University of Hiroshima\\
1-1-71, Ujina-Higashi, Minami-ku, \\
Hiroshima 734-8558, Japan\\
E-mail: skamada@pu-hiroshima.ac.jp}
\and
\IEEEauthorblockN{Takumi Ichimura}
\IEEEauthorblockA{Advanced Artificial Intelligence Project Research Center,\\
Research Organization of Regional Oriented Studies,\\
Faculty of Management and Information System,\\
Prefectural University of Hiroshima\\
1-1-71, Ujina-Higashi, Minami-ku, \\
Hiroshima 734-8558, Japan\\
E-mail: ichimura@pu-hiroshima.ac.jp}
\and
\IEEEauthorblockN{Takashi Iwasaki}
\IEEEauthorblockA{MCC Laboratory, Infrastructure System Group,\\
Mitsui Consultants Co. Ltd.\\
Gate City Ohsaki West Tower 15F,
1-11-1, Osaki, Shinagawa,\\
Tokyo 141-0032, Japan\\}
}

\maketitle

\begin{abstract}
  We have developed an adaptive structural Deep Belief Network (Adaptive DBN) that finds an optimal network structure in a self-organizing manner during learning. The Adaptive DBN is the hierarchical architecture where each layer employs Adaptive Restricted Boltzmann Machine (Adaptive RBM). The Adaptive RBM can find the appropriate number of hidden neurons during learning. The proposed method was applied to a concrete image benchmark data set SDNET2018 for crack detection. The dataset contains about 56,000 crack images for three types of concrete structures: bridge decks, walls, and paved roads. The fine-tuning method of the Adaptive DBN can show 99.7\%, 99.7\%, and 99.4\% classification accuracy for three types of structures. However, we found the database included some wrong annotated data which cannot be judged from images by human experts. This paper discusses consideration that purses the major factor for the wrong cases and the removal of the adversarial examples from the dataset.
\end{abstract}


\section{Introduction}

In recent years, Artificial Intelligence (AI) related technology has shown remarkable development centering on deep learning. With major achievements in image recognition and speech recognition, AI continues to disrupt society because recent deep learning has exploded with interesting and promising results. In addition to developing deep learning models with high originality, large amounts of data are collected to find regularity and relevance, and to make judgment and prediction by using the developed models in the real world problems.

The adaptive structural learning method of Deep Belief Network (Adaptive DBN) \cite{Kamada18_Springer} has an outstanding function to find the optimal network structure of Restricted Boltzmann Machine (RBM) \cite{Hinton06,Hinton12} which is the self-organizing process by hidden neuron generation and annihilation algorithm during learning phase. The algorithm observes the situation of training parameters at the RBM network. Adaptive DBN is the hierarchical model of RBMs where a new RBM is also automatically generated to monitor the total error of deep learning. Adaptive DBN method shows the highest classification capability for image recognition of some benchmark data sets such as MNIST \cite{LeCun98a}, CIFAR-10, and CIFAR-100 \cite{CIFAR10}. The paper \cite{Kamada18_Springer} reported the model can reach higher classification accuracy for test cases than existing Convolutional Neural Network (CNN) such as AlexNet \cite{AlexNet}, GoogLeNet \cite{GoogLeNet}, VGG16 \cite{VGG16}, and ResNet \cite{ResNet}.

According to a report by McKinsey \cite{McKinsey}, the civil construction sector has a net worth of more than \$10 trillion a year, the industry had been relatively under digitized until recently. Also in the field of civil engineering, AI related to technologies has more sophisticated applications in construction management, design optimization, risk control, and quality control. For example, the immediate maintenance and inspection operation for a road structure, bridge, debris barrier, and so on, was required after extraordinary weather such as torrential rain and typhoon. The inspection for concrete bridge is not easy operation, because the bridge inspection car should have special equipment such as ladder to watch the surface at a high altitude. Moreover, the experts for crack detection are decreasing. The deep learning technology for crack detection in the field of civil engineering is required to solve the problems.

SDNET2018 \cite{SDNET2018} is an annotated concrete image dataset for crack detection algorithms. It consists of over 56,000 cracked and non-cracked images for three types of structure; concrete bridge decks, walls, and pavements. They are used for training of crack detection algorithms such as deep learning \cite{CNN_CONCRETE}. The dataset includes images with many kinds of obstructions such as shadows, surface roughness, scaling, edges, holes, and background debris. SDNET2018 will be one of benchmark dataset in the field of structural health monitoring and the concrete crack detection method have been developed by using some deep learning models \cite{Cha2017, Zhang2016, Zhang2017}. In \cite{SDNET2018}, the classification accuracy was reported 91.9\%, 89.3\%, and 95.5\% for three types of structures as the results of transfer learning by using the AlexNet \cite{AlexNet}. On the other hand, our proposed method, Adaptive DBN, can reach 96.5\%, 96.8\%, and 96.5\% classification accuracy and the values are higher than the result reported in \cite{SDNET2018}. After training of Adaptive DBN, the fine-tuning method described in \cite{Kamada17_IJCIStudies} can show 99.7\%, 99.7\%, and 99.4\%, respectively. The fine-tuning method patches a part of network signal flow and it will be a helpful method to improve the classification capability. Accordingly the method can reduce the computational cost because of the inference by the explicit knowledge.

However, we found that the database included some wrong annotated data which cannot be judged from images by human experts. This paper discusses consideration that purses the major factor for the wrong cases and the removal of the adversarial data from dataset \cite{Goodfellow2015}. The fine-tuning method can train the network weights to follow the annotated label against the label is regardless of correct or wrong. In Japan, the concrete crack standards is given by the Japan Concrete Engineering Association. The difference of annotation is caused according to the standard for concrete cracks by country. However, the difference of annotation causes to be a recent big topic in the future of AI. The case can be the demonstration of adversarial examples as a known example of how fragile neural networks can be. For the problem, we investigate the open benchmark dataset SDNET2018 by experts and report the classification result that trains the dataset removing or modifying adversarial examples.

\section{Adaptive Structural Learning Method of Deep Belief Network}
\label{sec:Adaptive_dbn}

The basic idea of our proposed Adaptive DBN is described in this section to understand the basic behavior of self-organized structure briefly.

\subsection{RBM and DBN}
\label{sec:RBMDBN}
\begin{figure}[bt]
\centering
\includegraphics[scale=0.8]{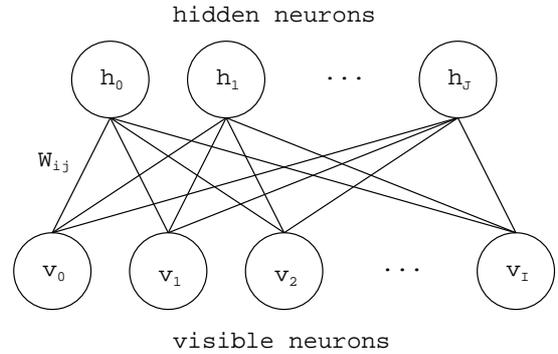}
%\vspace{-3mm}
\caption{Network structure of RBM}
\label{fig:rbm}
\end{figure}

A RBM \cite{Hinton12} is a kind of stochastic model for unsupervised learning. In the network structure of RBM, there are two kinds of binary layers as shown in Fig.~\ref{fig:rbm}: a visible layer $\bvec{v} \in \{0, 1 \}^{I}$ for input patterns and a hidden layer $\bvec{h} \in \{0, 1 \}^{J}$. The parameters $\bvec{b} \in \mathbb{R}^{I}$ and $\bvec{c} \in \mathbb{R}^{J}$ can work to make an adjustment of behaviors for a visible neuron and a hidden neuron, respectively. The weight $W_{ij}$ is the connection between a visible neuron $v_{i}$ and a hidden neuron $h_j$. A RBM minimizes the energy function $E(\bvec{v}, \bvec{h})$ during training in the following equations.
\begin{equation}
E(\bvec{v}, \bvec{h}) = - \sum_{i} b_i v_i - \sum_j c_j h_j - \sum_{i} \sum_{j} v_i W_{ij} h_j ,
\label{eq:energy}
\end{equation}
%\vspace{-2mm}
\begin{equation}
p(\bvec{v}, \bvec{h})=\frac{1}{Z} \exp(-E(\bvec{v}, \bvec{h})),
\label{eq:prob}
\end{equation}
%\vspace{-2mm}
\begin{equation}
Z = \sum_{\bvec{v}} \sum_{\bvec{h}} \exp(-E(\bvec{v}, \bvec{h})),
\label{eq:z}
\end{equation}
% 1 Dec, 19

Eq.~(\ref{eq:prob}) is a probability of $\exp(-E(\bvec{v}, \bvec{h}))$. $Z$ is calculated by summing energy for all possible pairs of visible and hidden vectors in Eq.~(\ref{eq:z}). The parameters $\bvec{\theta}=\{\bvec{b}, \bvec{c}, \bvec{W} \}$ is optimized by partial derivative of $p(\bvec{v})$ for given input data $\bvec{v}$ . A standard estimation for the parameters is maximum likelihood in a statistical model $p(\bvec{v})$.

DBN \cite{Hinton06} is a stacking box for stochastic unsupervised learning by hierarchically organized several pre-trained RBMs. The conditional probability of a the $j$-th hidden neuron at the $l$-th RBM is defined by Eq.~(\ref{eq:prob_dbn}).

\begin{equation}
\label{eq:prob_dbn}
p(h_j^{l} = 1 | \bvec{h}^{l-1})= sigmoid(c^{l}_j + \sum_{i}W^{l}_{ij} h^{l-1}_{i}),
\end{equation}
where $c^{l}_j$ and $W^{l}_{ij}$ indicate the parameters for the $j$-th hidden neuron and the weight at the $l$-th RBM, respectively. $\bvec{h}^{0} = \bvec{v}$ means the given input data. In order to make the DBN a supervised learning for a classification task, the last layer appends to the final output layer to calculate the output probability $y_k$ for a category $k$. The calculation is implemented by Softmax in Eq.(\ref{eq:softmax}). 
\begin{equation}
\label{eq:softmax}
y_k = \frac{\exp(z_{k})}{\sum^{M}_{j} \exp(z_j)},
\end{equation}
where $z_{j}$ is an output pattern of the $j$-th hidden neuron at the output layer. The number of output neurons is $M$. The difference between the output $y_k$ and the teacher signal for the category $k$ is minimized. 

\subsection{Neuron Generation and Annihilation Algorithm of RBM}
\label{subsec:adaptive_rbm}
Recently, deep learning models have higher classification capability for given large amount of data, however the size of its network structure or the number of its parameters that a network designer determines must become larger. For such a problem, the adaptive structural learning method in RBM, Adaptive RBM \cite{Kamada18_Springer}, was developed to find the optimal number of hidden neurons for given input space during its training situation. The key idea of Adaptive RBM is the observation of the variance of parameters called Walking Distance (WD) \cite{Kamada16_SMC}. If there are not enough neurons to classify the input images in the RBM, then WD may fluctuate even after a long training. According to \cite{Kamada18_Springer}, we selected to monitor two parameters ($\bvec{c}$ and $\bvec{W}$) of $\bvec{\theta}=\{\bvec{b}, \bvec{c}, \bvec{W} \}$ which have influenced on the learning convergence of RBM except the parameter, $\bvec{b}$, related to input data. 

\begin{figure}[]
\begin{center}
\subfigure[Neuron generation]{\includegraphics[scale=0.5]{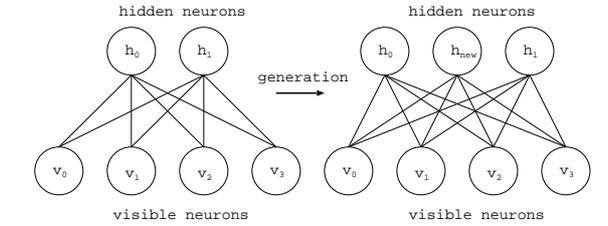}\label{fig:neuron_generation}}
\subfigure[Neuron annihilation]{\includegraphics[scale=0.5]{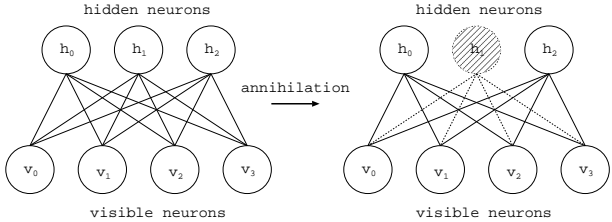}\label{fig:neuron_annihilation}}
\vspace{-3mm}
\caption{Adaptive RBM}
\label{fig:adaptive_rbm}
\end{center}
\end{figure}

However, we may meet a situation that some redundant neurons were generated due to the neuron generation process. The neuron annihilation algorithm was operated to kill the corresponding neuron after neuron generation process. Fig.~\ref{fig:neuron_annihilation} shows that the corresponding neuron is annihilated \cite{Kamada16_ICONIP}.

\begin{figure*}[]
\centering
\includegraphics[scale=0.8]{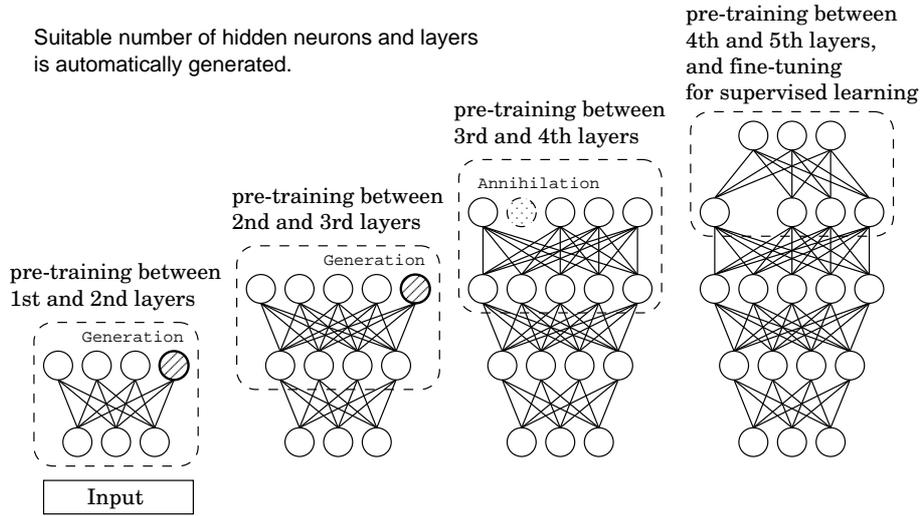}
\vspace{-3mm}
\caption{An Overview of Adaptive DBN}
\label{fig:adaptive_dbn}
\end{figure*}

\subsection{Layer Generation Algorithm of DBN}
\label{subsec:adaptive_dbn}
Fig.~\ref{fig:adaptive_dbn} shows the overview of layer generation in Adaptive DBN. Our DBN model is a generative neural network model with hierarchical layered of the two or more pre-trained RBMs. The input of $l+1$-th RBM is given by the output of $l$-th RBM. DBN with many layered RBMs will have higher classification capability of data representation. Such hierarchical model can represent the various features from the lower-level features to concrete image in the direction of input layer to output layer. In order to realize the sophisticated learning, the optimal number of RBMs depends on the target data space \cite{Kamada16_TENCON}.

Our developed Adaptive DBN can accomplish to make an optimal network structure to automatically add the RBM one by one based on the idea of WD described in section \ref{subsec:adaptive_rbm}. If the energy values and WD don't converge, each RBM lacks presentation capability to classify input patterns. In order to solve such a situation, the refined network structure is self-organized by a new generated RBM to obtain higher classification power. Eqs.~(\ref{eq:layer_generation1})-(\ref{eq:layer_generation2}) are the condition for layer generation is defined by using the WD of RBM and the energy function as follows.

\begin{equation}
\sum_{l=1}^{k} WD^{l} > \theta_{L1},
\label{eq:layer_generation1}
\vspace{-3mm}
\end{equation}
\begin{equation}
\sum_{l=1}^{k} E^{l} > \theta_{L2}
\label{eq:layer_generation2}
%\vspace{-3mm}
\end{equation}

\section{SDNET2018}
\label{sec:SDNET2018}
\subsection{Data Description}
\label{sec:Data_Description}
\begin{figure}[btp]
  \centering
  \subfigure[Bridge deck, w/o cracks]{\includegraphics[scale=0.3]{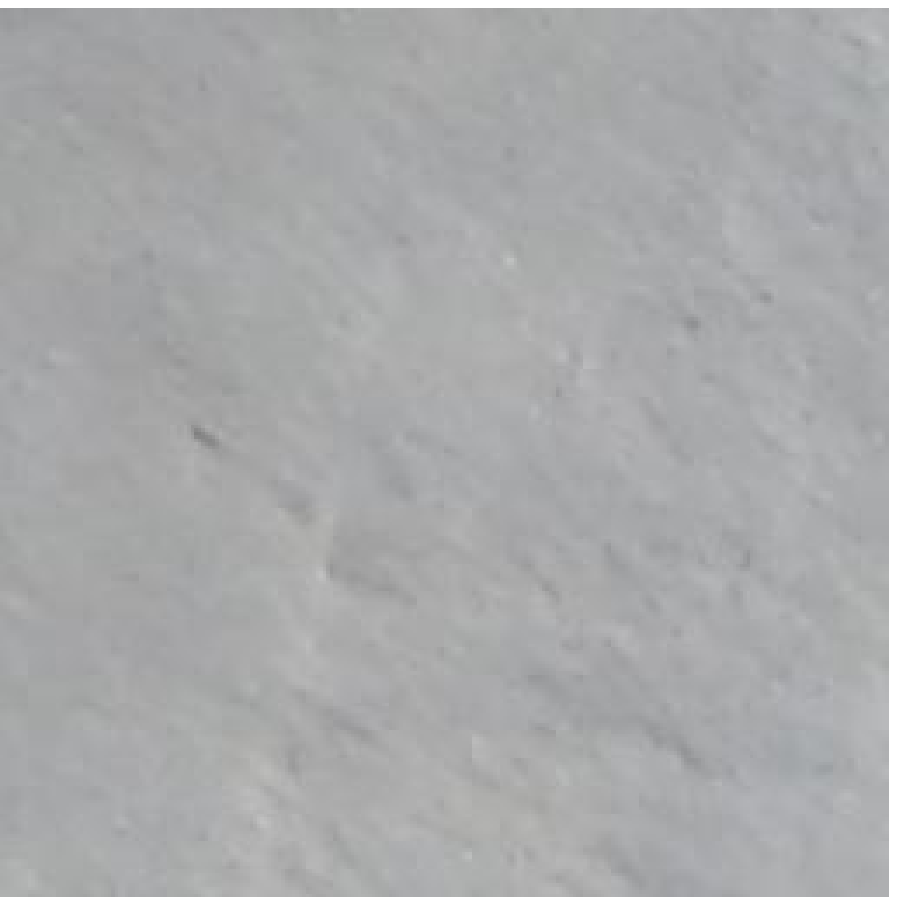}\label{fig:UD}}
  \subfigure[Bridge deck, with cracks]{\includegraphics[scale=0.3]{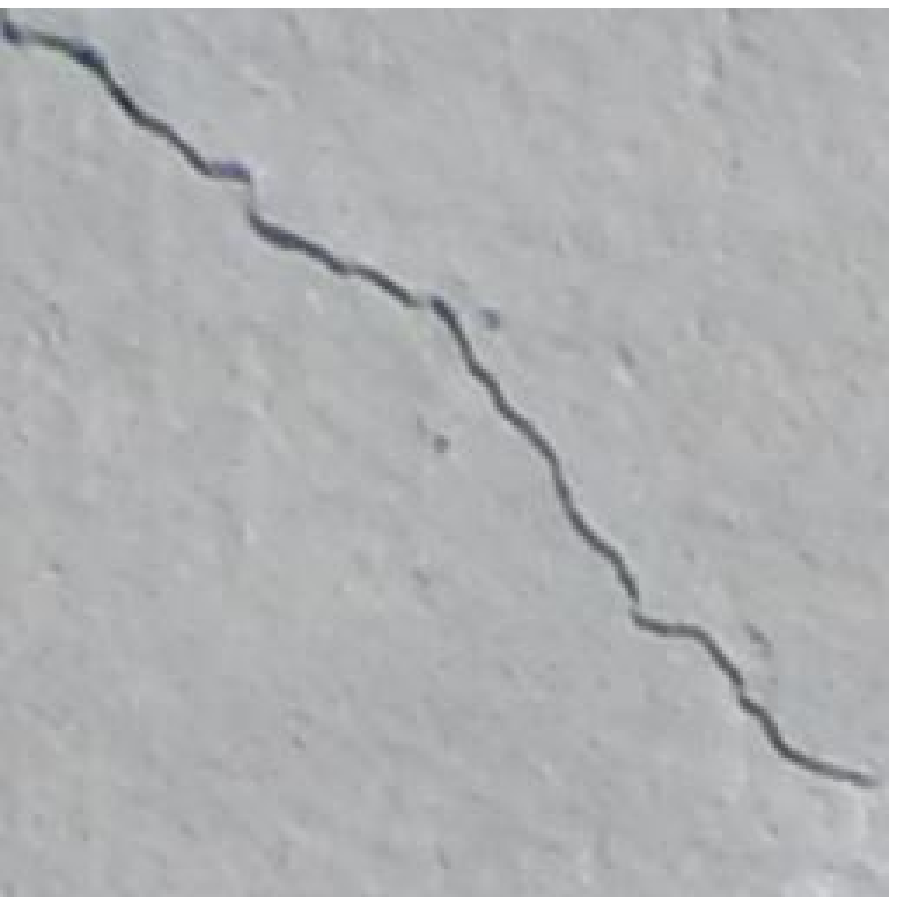}\label{fig:CD}}
  \subfigure[Wall, w/o cracks]{\includegraphics[scale=0.3]{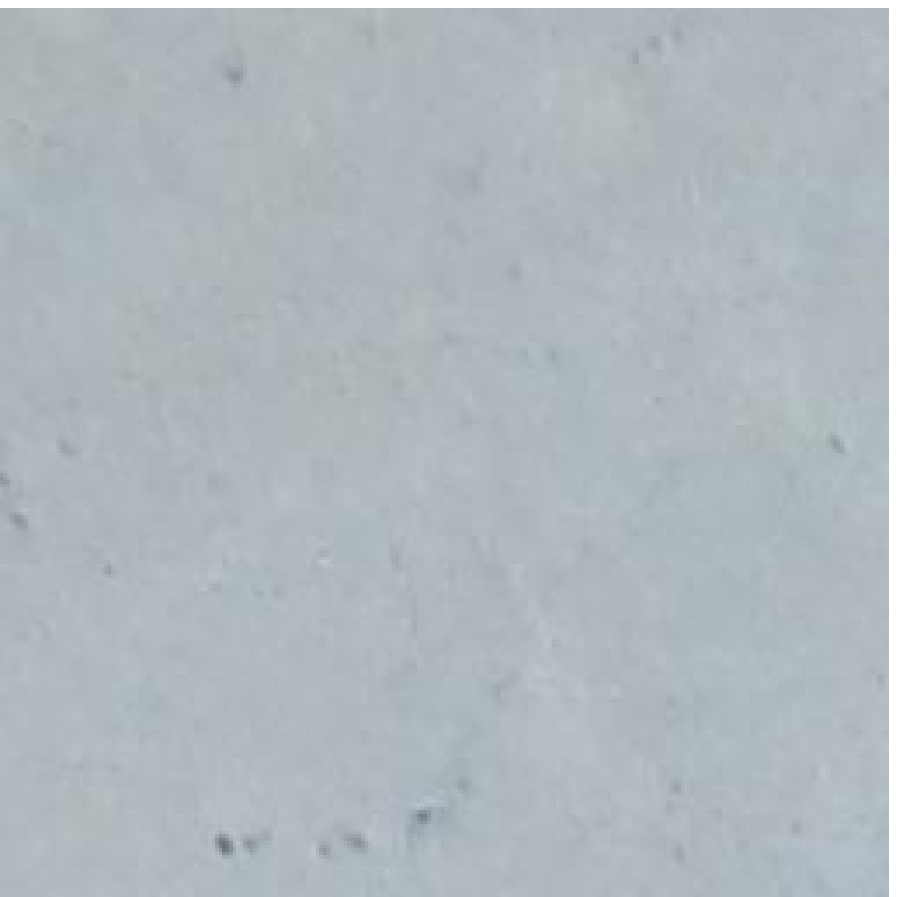}\label{fig:UW}}
  \subfigure[Wall, with cracks]{\includegraphics[scale=0.3]{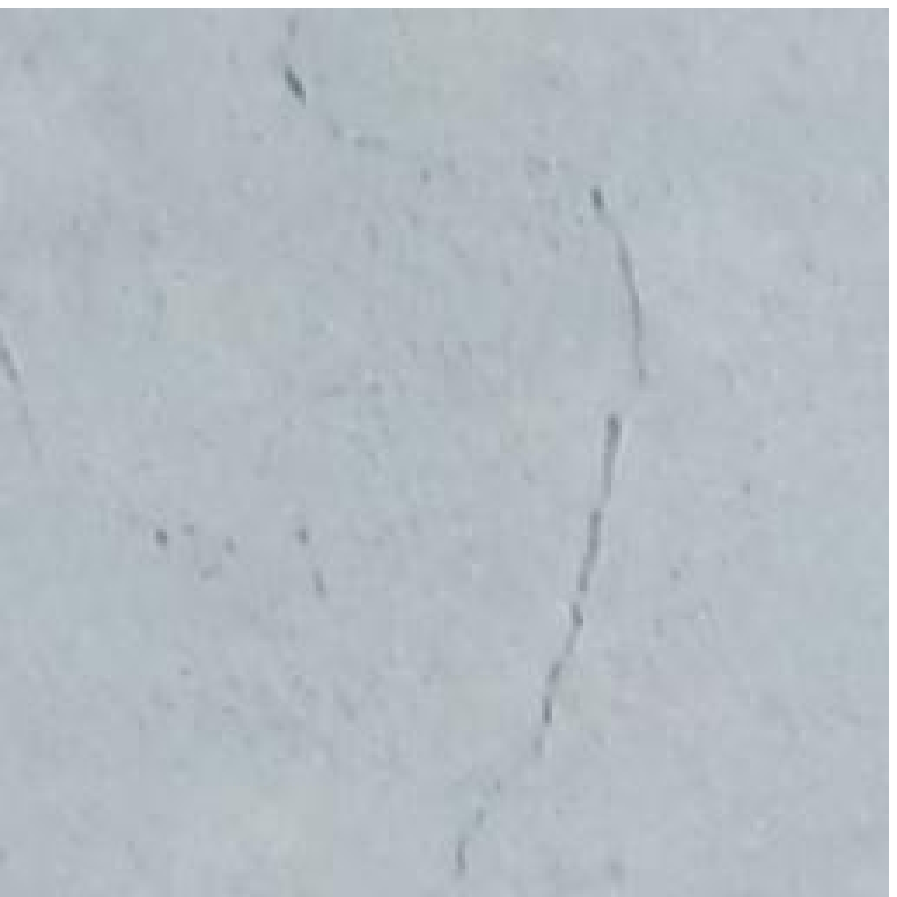}\label{fig:CW}}
  \subfigure[Pavement, w/o cracks]{\includegraphics[scale=0.3]{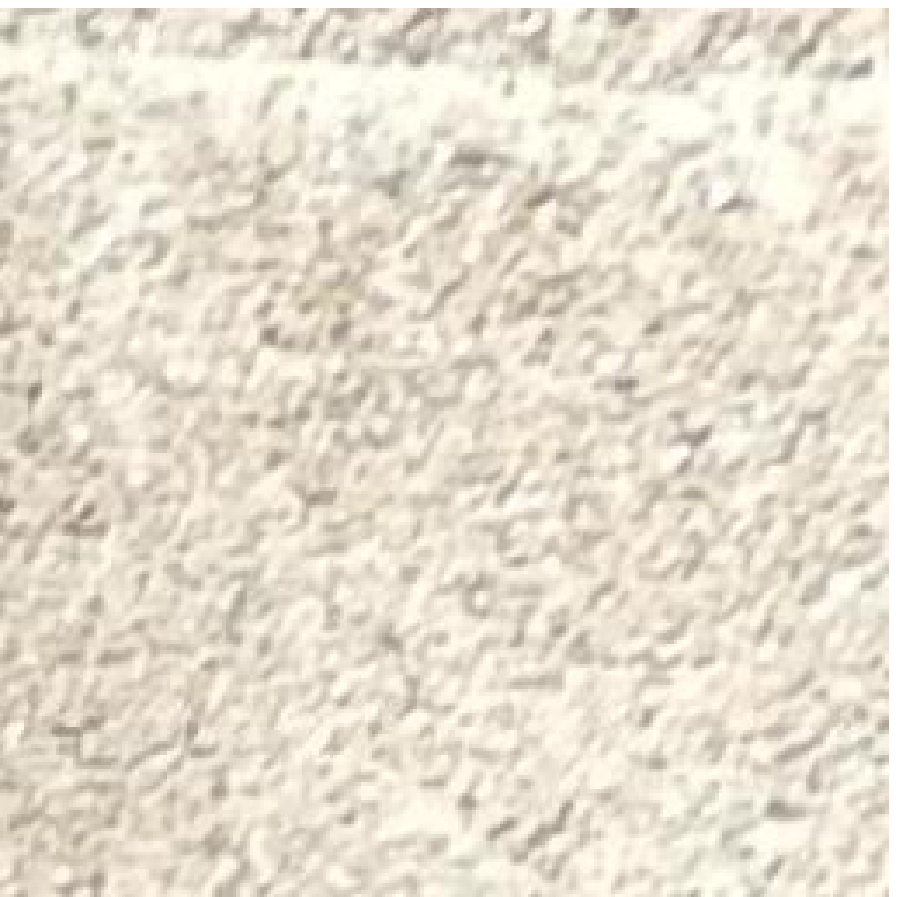}\label{fig:UP}}
  \subfigure[Pavement, with cracks]{\includegraphics[scale=0.3]{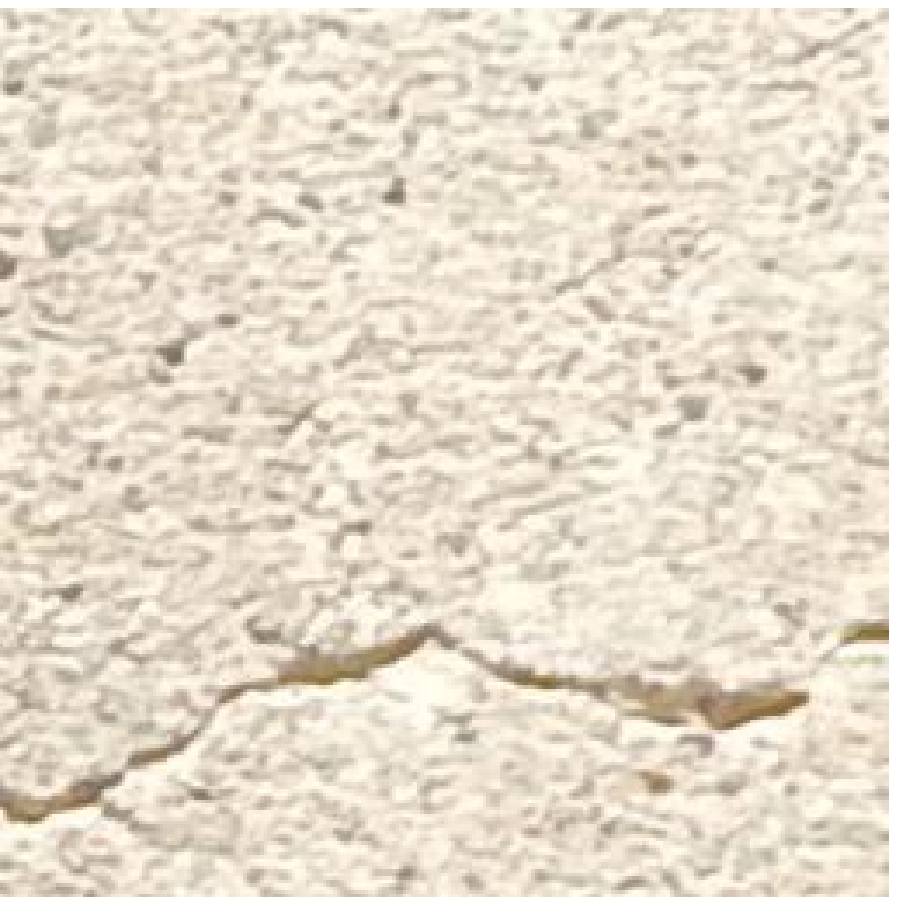}\label{fig:CP}}
  \caption{Sample of SDNET2018}
  \label{fig:data_sdnet_sample}
%\vspace{-5mm}
\end{figure}

SDNET2018 \cite{SDNET2018} is an annotated concrete image dataset collected from Utah State University. It consist of over 56,000 cracked and non-cracked images for three types of structure; bridge decks, walls, and pavements, as a classification problem. They are used to develop crack detection algorithm such as deep learning \cite{CNN_CONCRETE}. The dataset includes images with many kinds of obstructions such as shadows, surface roughness, scaling, edges, holes, and background debris. Table \ref{tab:data_sdnet_category} shows the number of cracked, non-cracked, and total sub-images of each type in SDNET2018. According to the report in \cite{SDNET2018}, the dataset includes cracks at the variety size from 0.06 mm to 25 mm. Fig.~\ref{fig:data_sdnet_sample} shows the sample images of crack conditions. Each image is 256 $\times$ 256 size in colored pixel and is annotated as `with cracked' or `without cracked' for the supervised training.

\begin{table}[btp]
\caption{The category of SDNET2018}
\vspace{-3mm}
\label{tab:data_sdnet_category}
\begin{center}
\scalebox{0.9}[0.9]{
\begin{tabular}{l|r|r}
\hline 
\multicolumn{1}{c|}{Category} & Training dataset & Test dataset  \\ 
\hline
Bridge deck w/o cracks   &  10,424  & 1,834 \\ 
Bridge deck with cracks  &     1,171  &    191 \\ \hline
Wall w/o cracks  &  12,853 & 1,434 \\ 
Wall with cracks  &    3,471 &     380 \\ \hline
Pavement w/o cracks  &  19,531  & 2,195 \\ 
Pavement with cracks  &     2,369  &    239 \\ \hline
\multicolumn{1}{c|}{total}  &     49,819  &    6,273 \\ \hline
\end{tabular}
} 
\end{center}
%\vspace{-5mm}
\end{table}

\subsection{Classification Results}
\label{sec:classification_results}
Table~\ref{tab:classification_ratio1} compares classification accuracy for SDNET test data with an existing CNN and the Adaptive DBN. The CNN is transfer learning based on AlexNet \cite{SDNET2018}. Our method can classify the images to cracked and non-cracked labels with accuracy of more than 95\% for three types of structures, which is higher than the existing CNN. Fine Tuning \cite{Kamada17_IJCIStudies} is a method to improve the classification accuracy of Adaptive DBN after learning. According to frequency of input/output patterns at each layer in the network after learning, the network weights are modified so that cases classified incorrectly are classified correctly. Fine Tuning method improved the classification accuracy for test data to more than 99.4\%.

Table~\ref{tab:classification_ratio2} shows the classification accuracy by Adaptive DBN as shown in Table~\ref{tab:classification_ratio1} for the training data and the test data for three types of structures without cracks and with cracks. The value in the parenthesis in the cell of the test data indicates ``the number of incorrect data / the total number of data''. The Adaptive DBN showed 100\% classification accuracy for training data. The test data showed classification accuracy of 95.3\% or more for all categories. The numerical values for the three types of structures were almost the same, and there was no significant difference. The numerical values did not drop significantly both with and without cracks.

\begin{table}[btp]
  \caption{Classification ratio of \cite{SDNET2018} and Adaptive DBN}
\vspace{-3mm}
\label{tab:classification_ratio1}
\begin{center}
\scalebox{0.9}[0.9]{
\begin{tabular}{l|r|r|r}
\hline
\multicolumn{1}{c|}{Category} &  \multicolumn{1}{c|}{CNN \cite{SDNET2018}} & \multicolumn{1}{c|}{Adaptive DBN} & \multicolumn{1}{c}{Fine Tuning \cite{Kamada17_IJCIStudies}} \\ \hline
Bridge deck &  91.9\%   & 96.5\% & 99.7\% \\ 
Wall              & 89.3\%  & 96.8\% & 99.7\% \\ 
Pavement   &  95.5\%  & 96.5\% & 99.4\%\\
\hline
\end{tabular}
} 
\end{center}
\vspace{-5mm}
\end{table}

\begin{table}[btp]
  \caption{Classification ratio by Adaptive DBN}
\vspace{-3mm}
\label{tab:classification_ratio2}
\begin{center}
\scalebox{0.9}[0.9]{
\begin{tabular}{l|r|rr}
\hline 
\multicolumn{1}{c|}{Category} & Train & \multicolumn{2}{c}{Test}  \\ 
\hline
Bridge deck w/o cracks    &  99.3\%  & 96.5\% & (64/1834)\\ 
Bridge deck with cracks   &  99.2\% & 96.3\% & (7/191) \\ \hline
Wall w/o cracks           & 98.8\%  & 97.2\% & (40/1434) \\ 
Wall with cracks          &  98.6\%  & 95.3\% & (18/380) \\ \hline
Pavement w/o cracks       &  98.2\%  & 96.6\% &  (75/2195) \\ 
Pavement with cracks      &  98.3\%  & 95.8\% &  (10/239) \\ \hline
\end{tabular}
} 
\end{center}
\vspace{-5mm}
\end{table}

\begin{figure}[bt]
\centering    
\subfigure[Sample 1-1]{\includegraphics[scale=0.35]{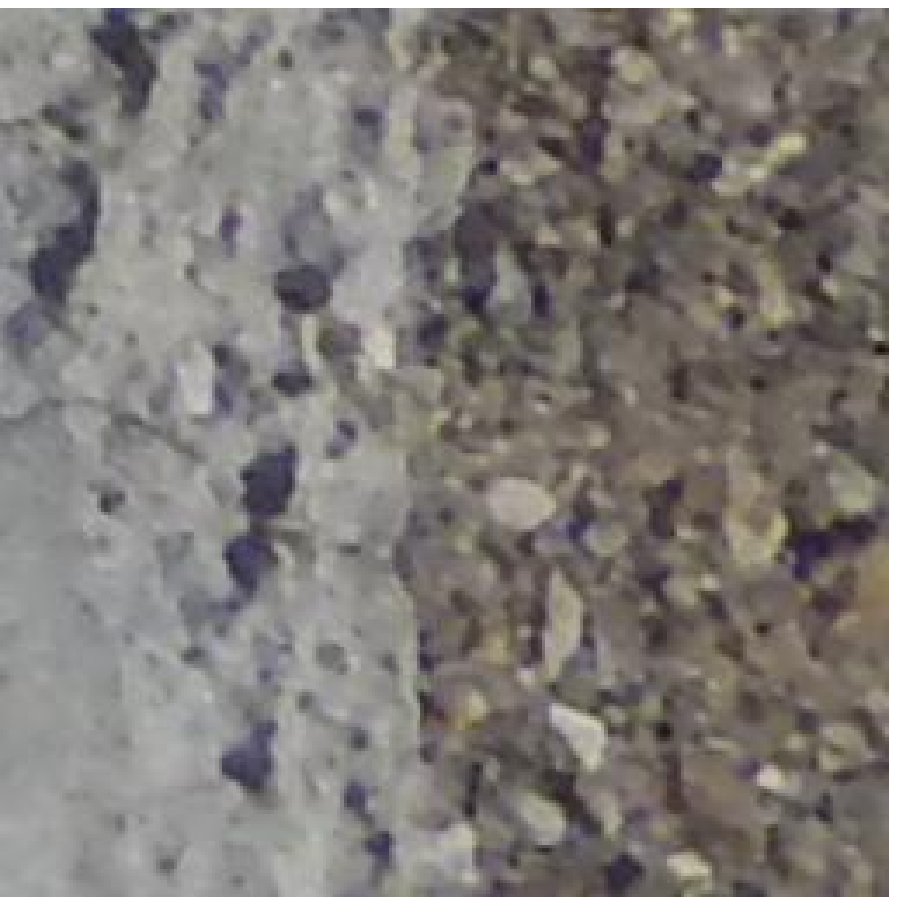}\label{fig:fig/error/CD/7046-198.eps}}
\subfigure[Sample 1-2]{\includegraphics[scale=0.35]{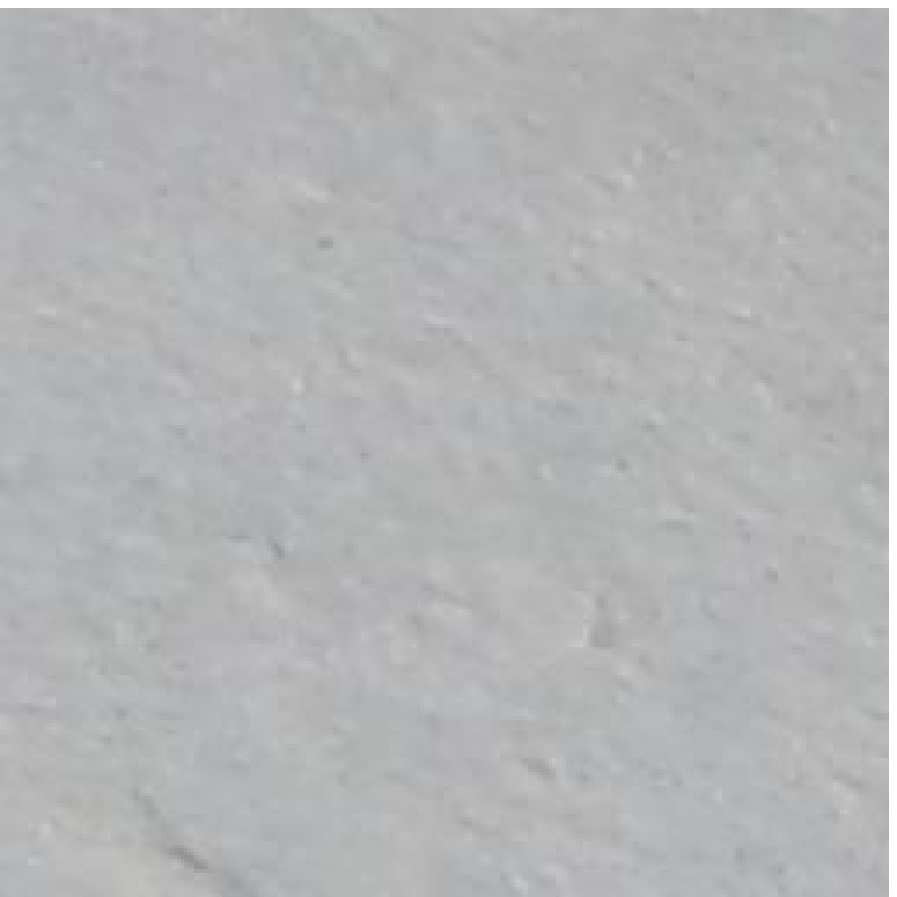}\label{fig:fig/error/CD/7009-73.eps}}
\subfigure[Sample 1-3]{\includegraphics[scale=0.35]{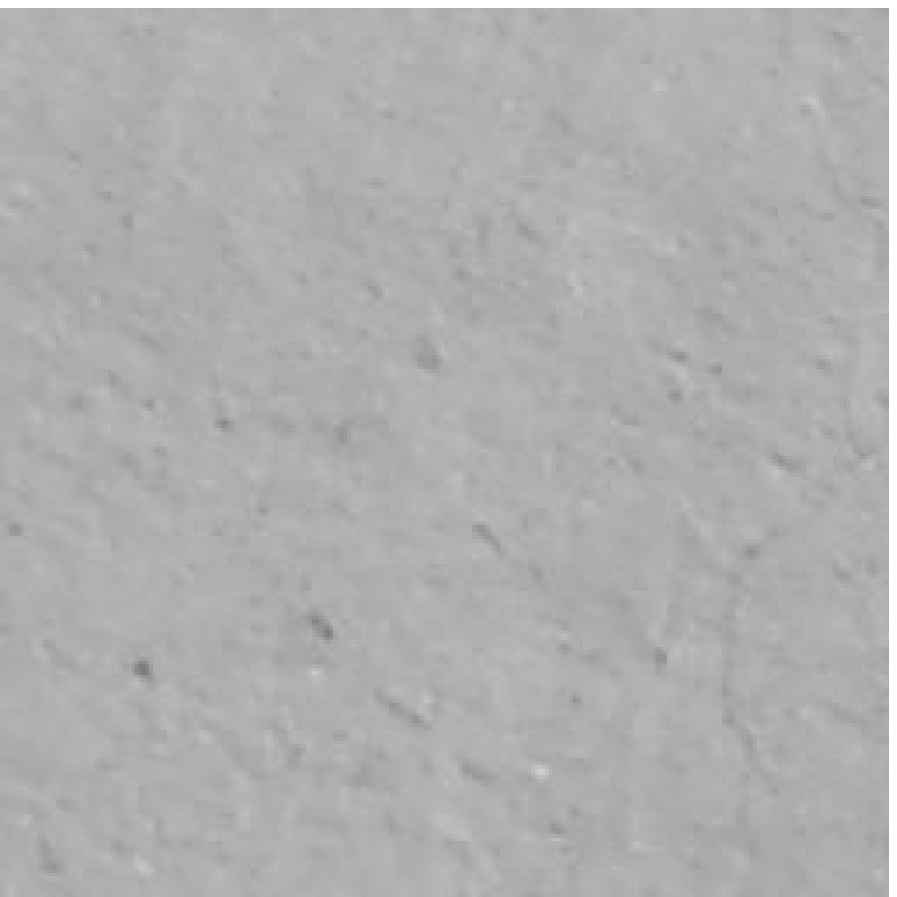}\label{fig:fig/error/CD/7001-56.eps}}
\subfigure[Sample 1-4]{\includegraphics[scale=0.35]{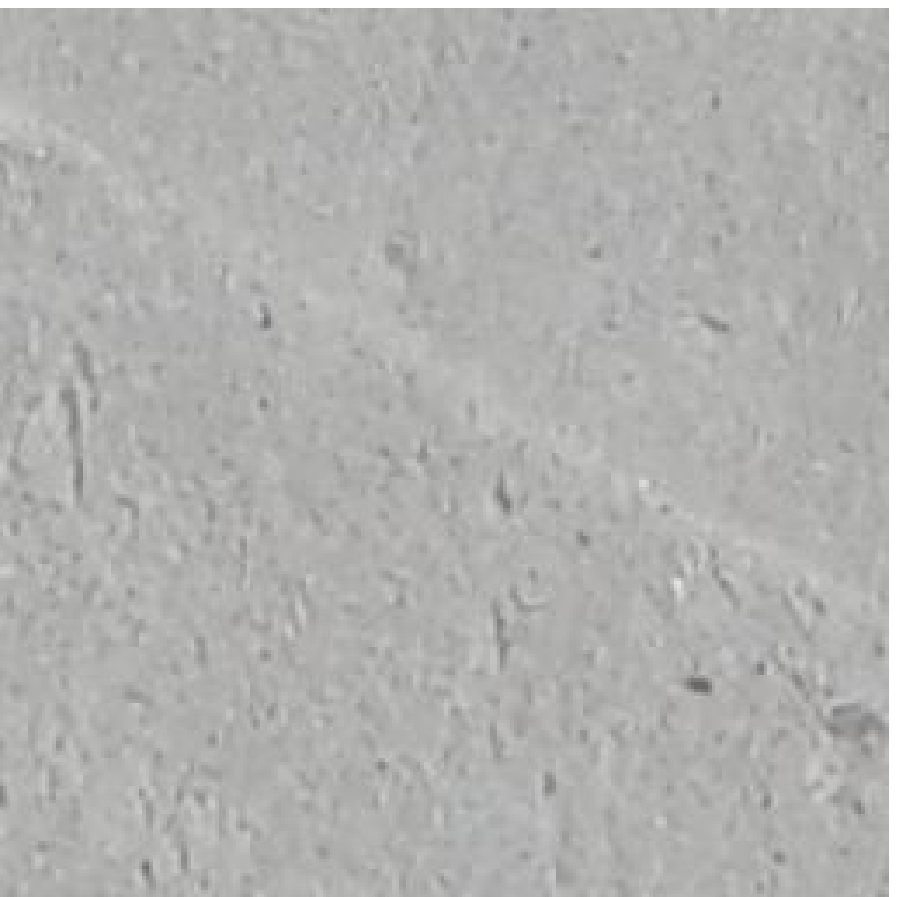}\label{fig:fig/error/CD/7012-172.eps}}
\vspace{-2mm}
\caption{1) Bridge deck, Mis-judgment for cracked images}
\label{fig:case_CD_UD}
\vspace{-3mm}
\end{figure}

\begin{figure}[bt]
\centering
\subfigure[Sample 2-1]{\includegraphics[scale=0.35]{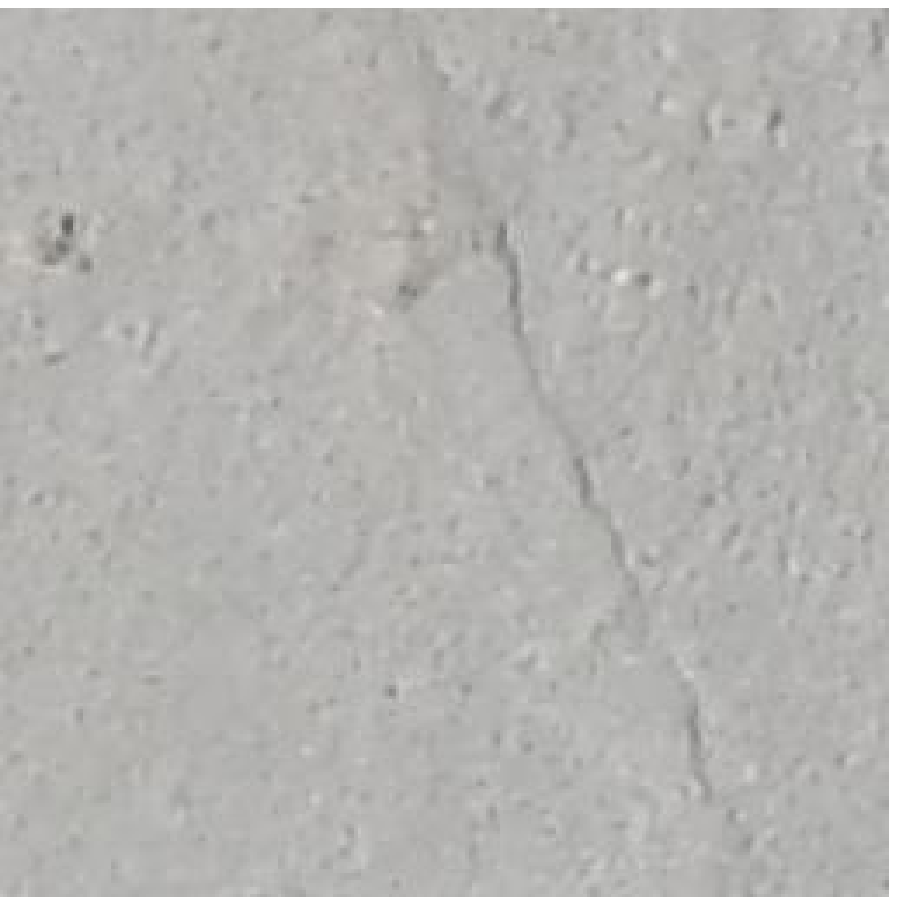}\label{fig:fig/error/UD/7043-130.eps}}
\subfigure[Sample 2-2]{\includegraphics[scale=0.35]{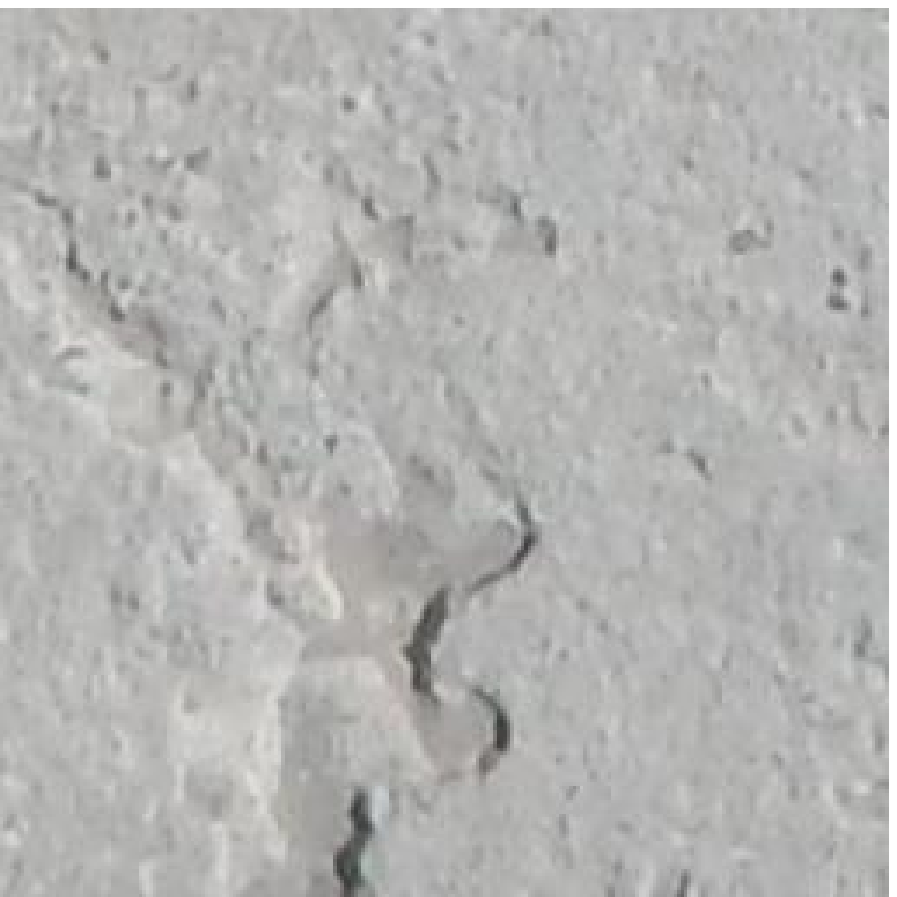}\label{fig:fig/error/UD/7003-225.eps}}
\subfigure[Sample 2-3]{\includegraphics[scale=0.35]{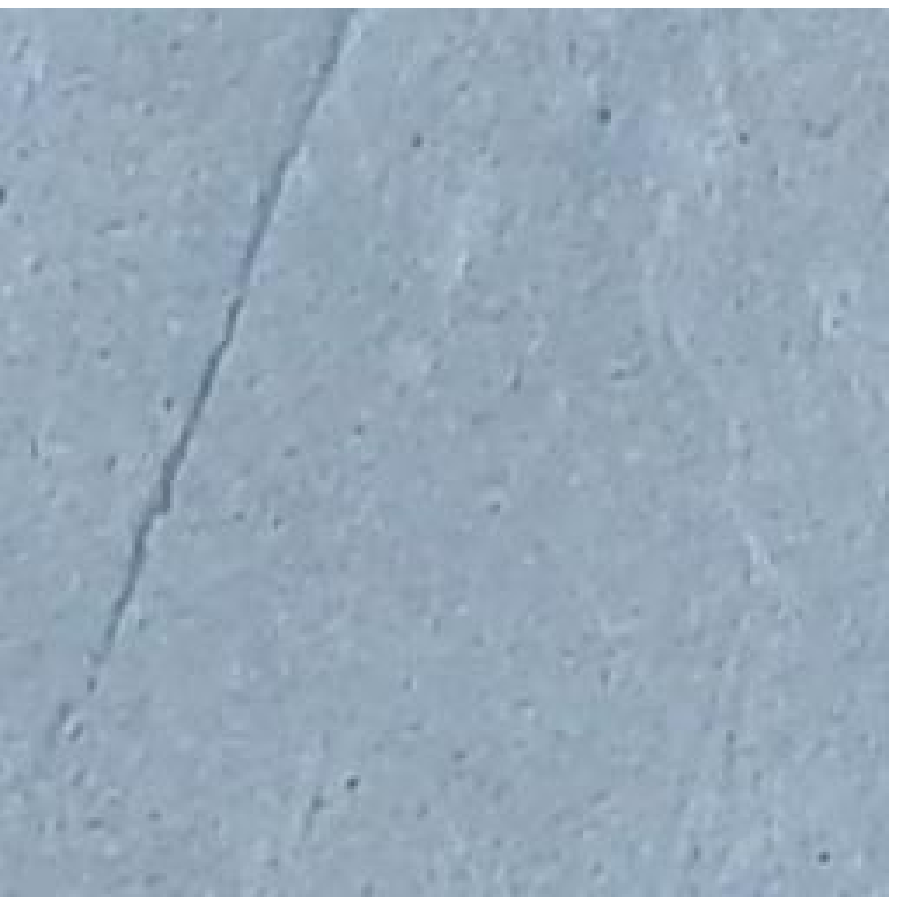}\label{fig:fig/error/UD/7057-187.eps}}
\subfigure[Sample 2-4]{\includegraphics[scale=0.35]{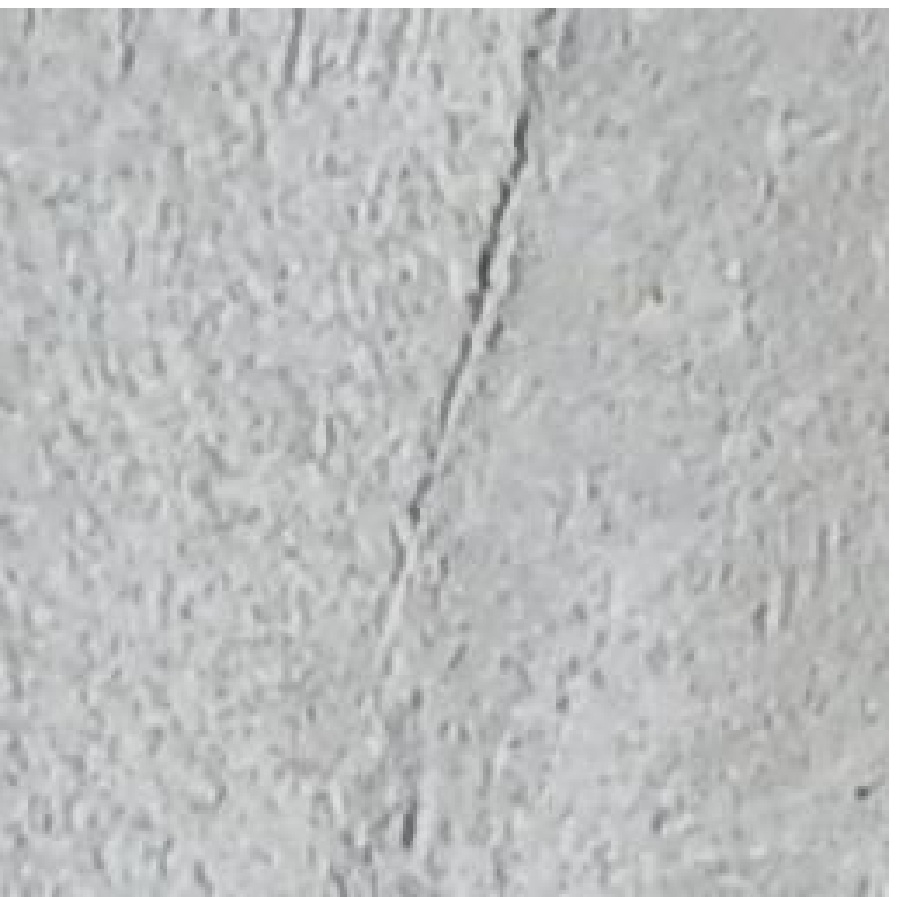}\label{fig:fig/error/UD/7054-63.eps}}
\vspace{-2mm}
\caption{2) Bridge deck, Mis-judgment for images without cracks}
\label{fig:case_UD_CD_1}
\vspace{-3mm}
\end{figure}

\begin{figure}[bt]
\centering
\subfigure[Sample 3-1]{\includegraphics[scale=0.35]{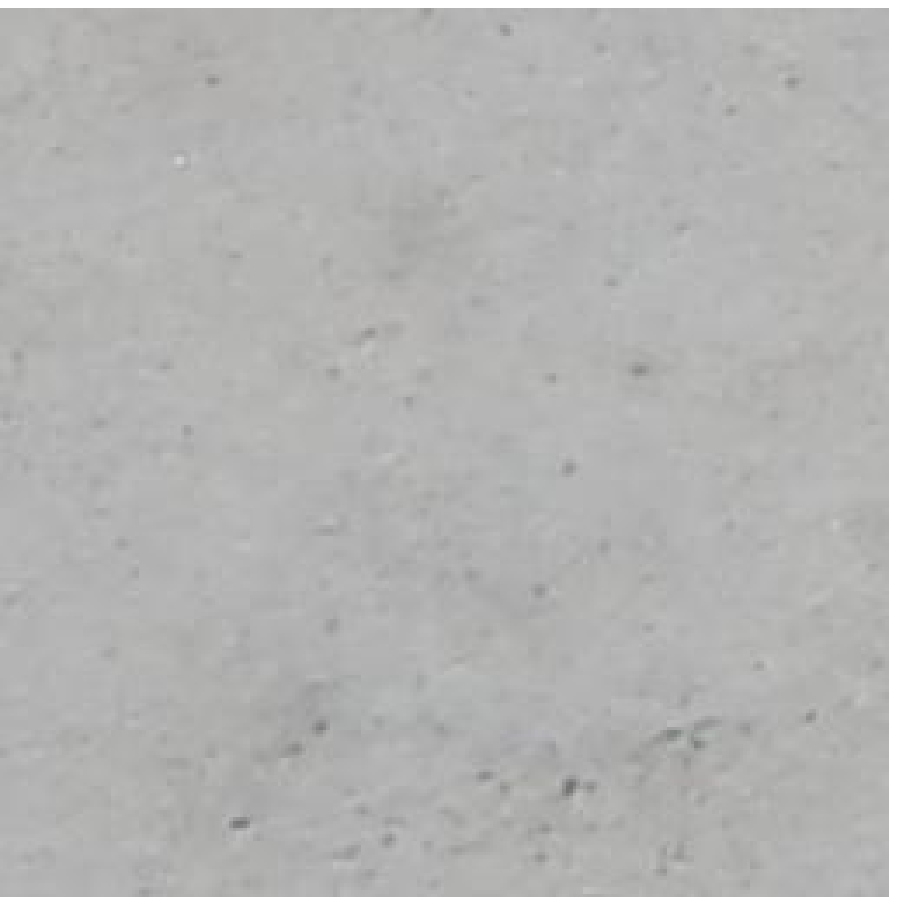}\label{fig:fig/error/CW/7093-165.eps}}
\subfigure[Sample 3-2]{\includegraphics[scale=0.35]{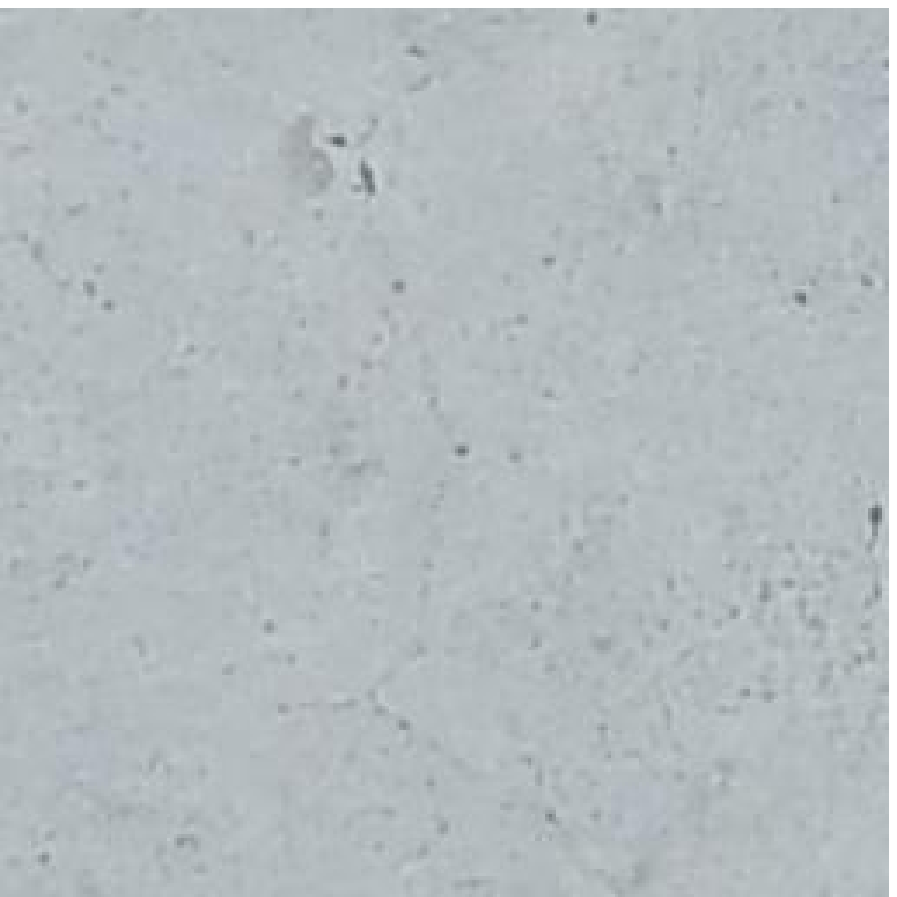}\label{fig:fig/error/CW/7069-151.eps}}
\subfigure[Sample 3-3]{\includegraphics[scale=0.35]{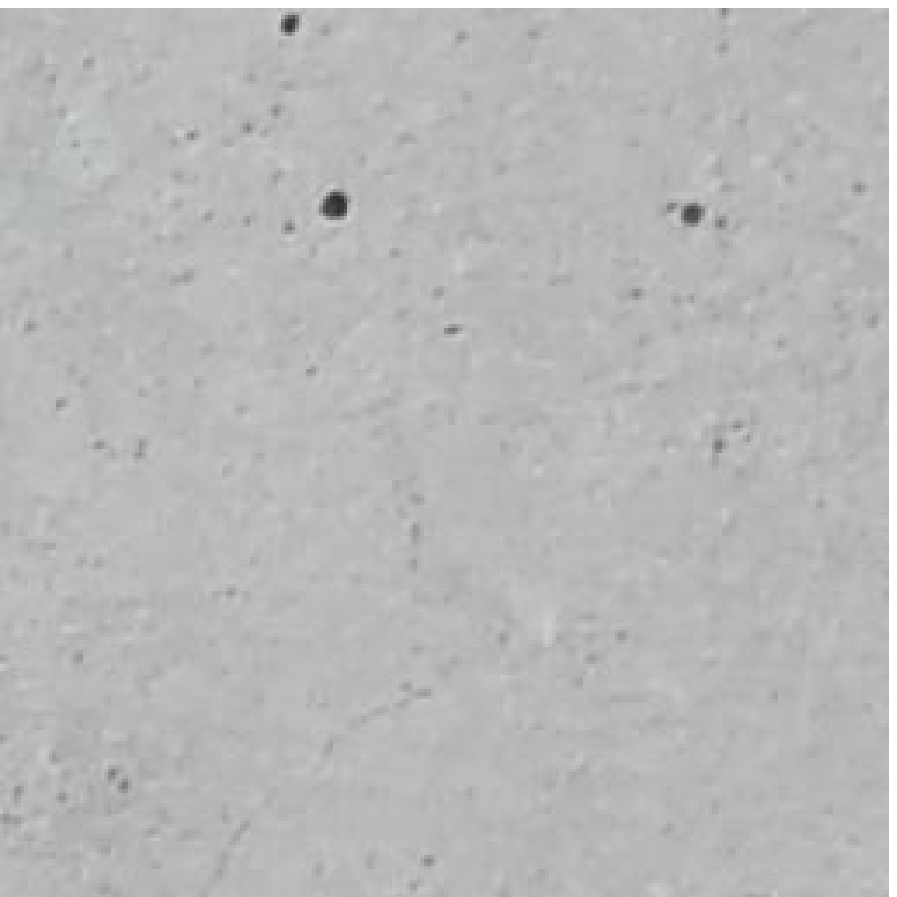}\label{fig:fig/error/CW/7072-85.eps}}
\subfigure[Sample 3-4]{\includegraphics[scale=0.35]{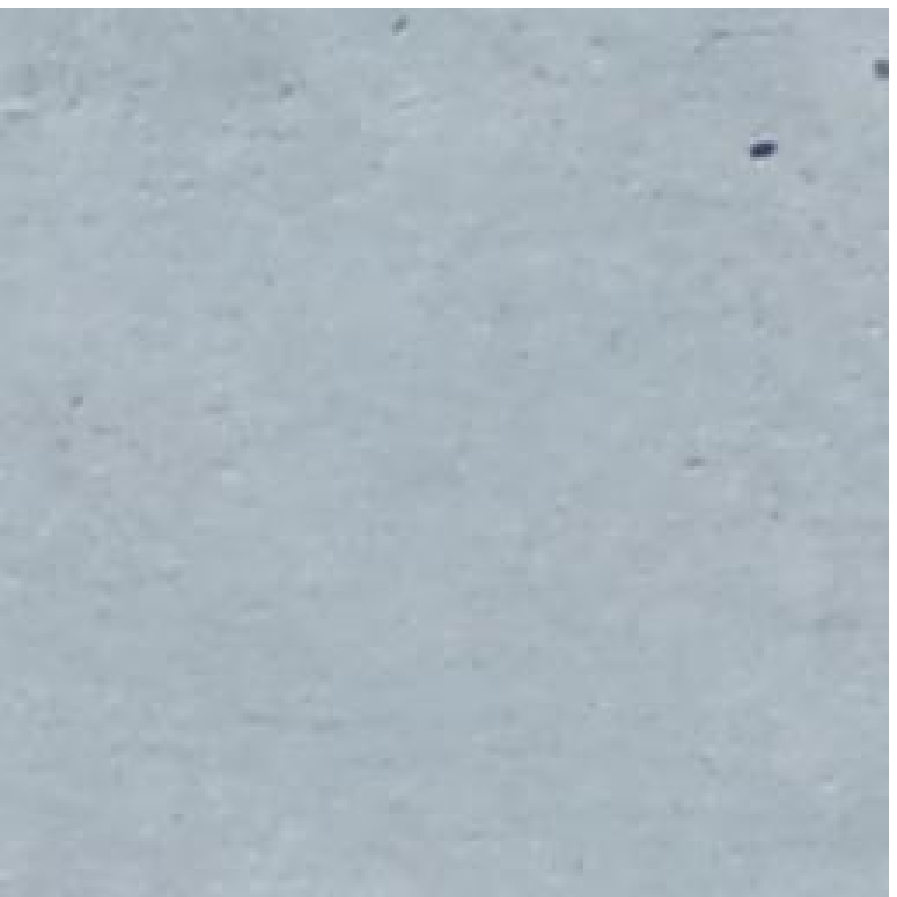}\label{fig:fig/error/CW/7086-107.eps}}
\vspace{-2mm}
\caption{3) Wall, Mis-judgment for cracked images}
\label{fig:case_CW_UW}
\vspace{-3mm}
\end{figure}

\begin{figure}[bt]
\centering
\subfigure[Sample 4-1]{\includegraphics[scale=0.35]{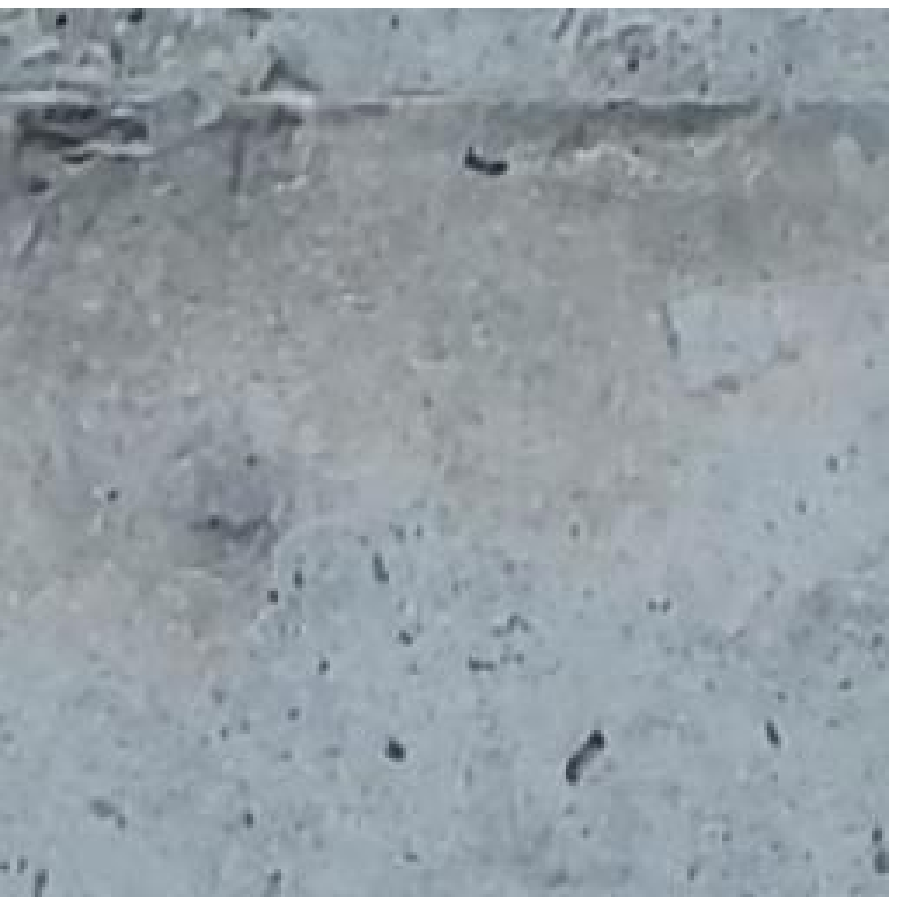}\label{fig:fig/error/UW/7130-152.eps}}
\subfigure[Sample 4-2]{\includegraphics[scale=0.35]{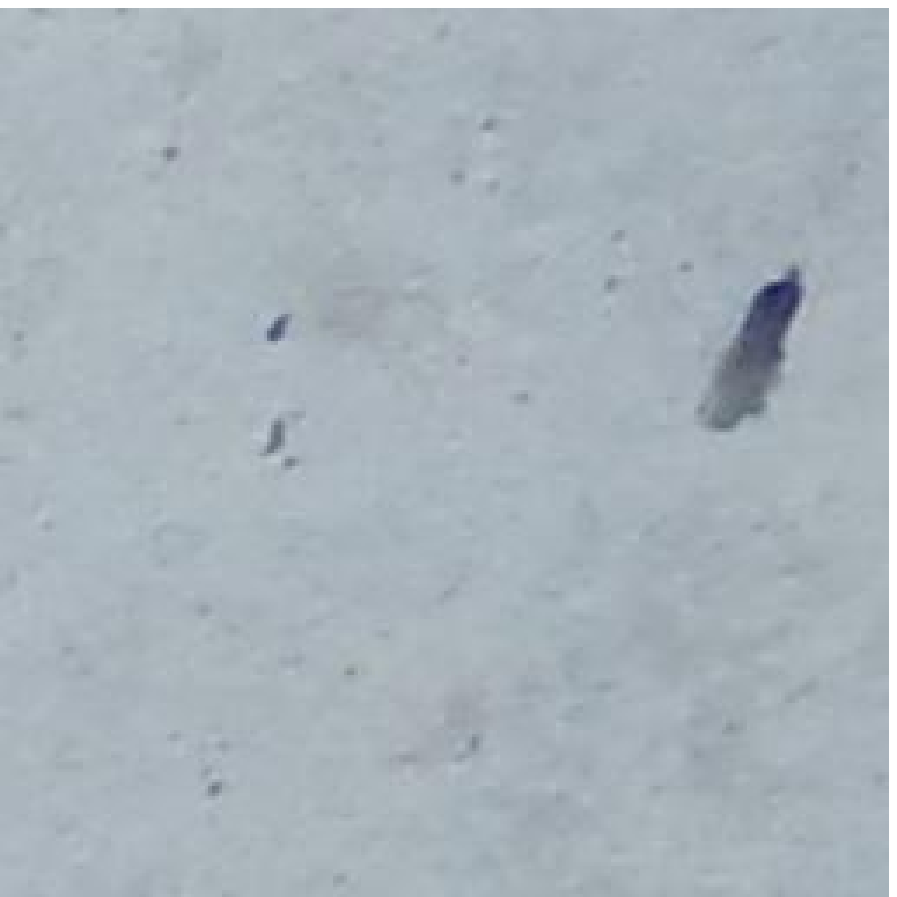}\label{fig:fig/error/UW/7078-90.eps}}
\subfigure[Sample 4-3]{\includegraphics[scale=0.35]{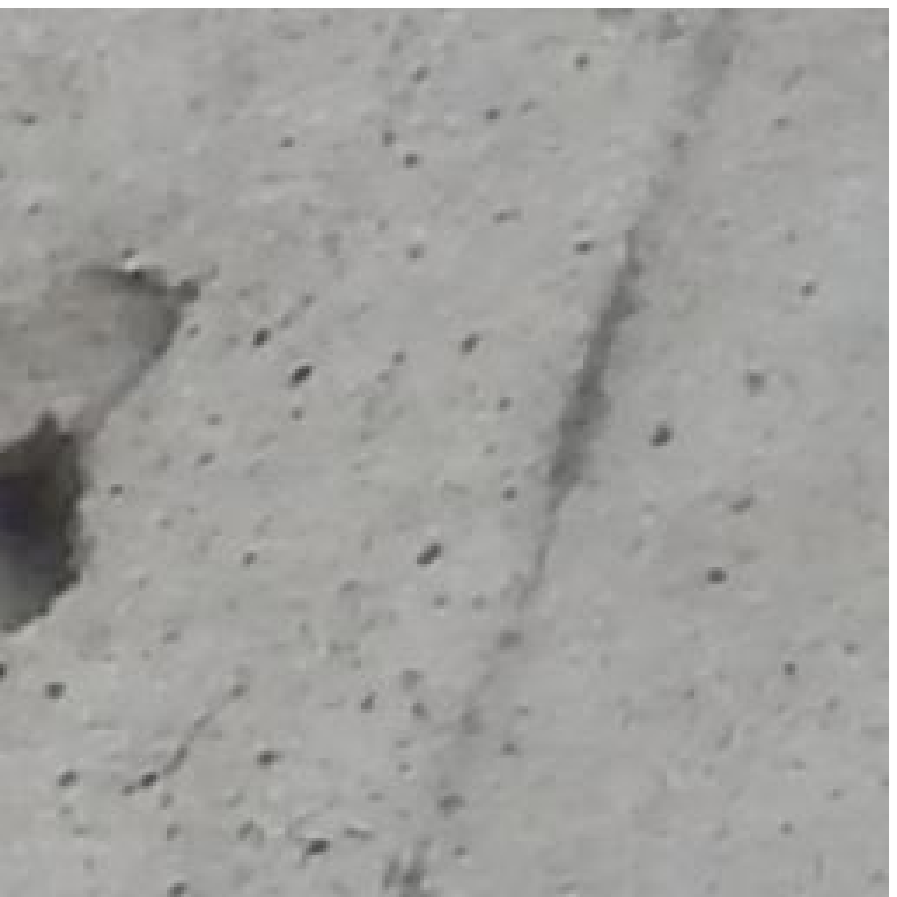}\label{fig:fig/error/UW/7094-89.eps}}
\subfigure[Sample 4-4]{\includegraphics[scale=0.35]{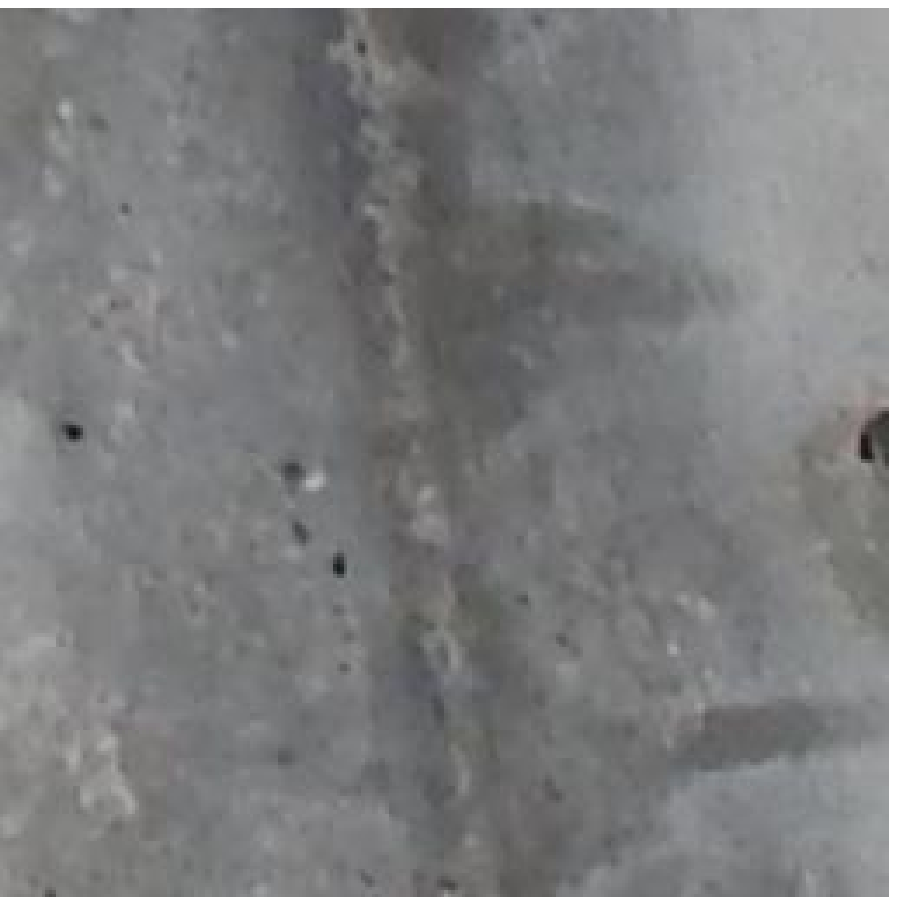}\label{fig:fig/error/UW/7112-42.eps}}
\vspace{-2mm}
\caption{4) Wall, Mis-judgment for images without cracks}
\label{fig:case_UW_CW_1}
\vspace{-3mm}
\end{figure}

\begin{figure}[bt]
\centering
\subfigure[Sample 5-1]{\includegraphics[scale=0.35]{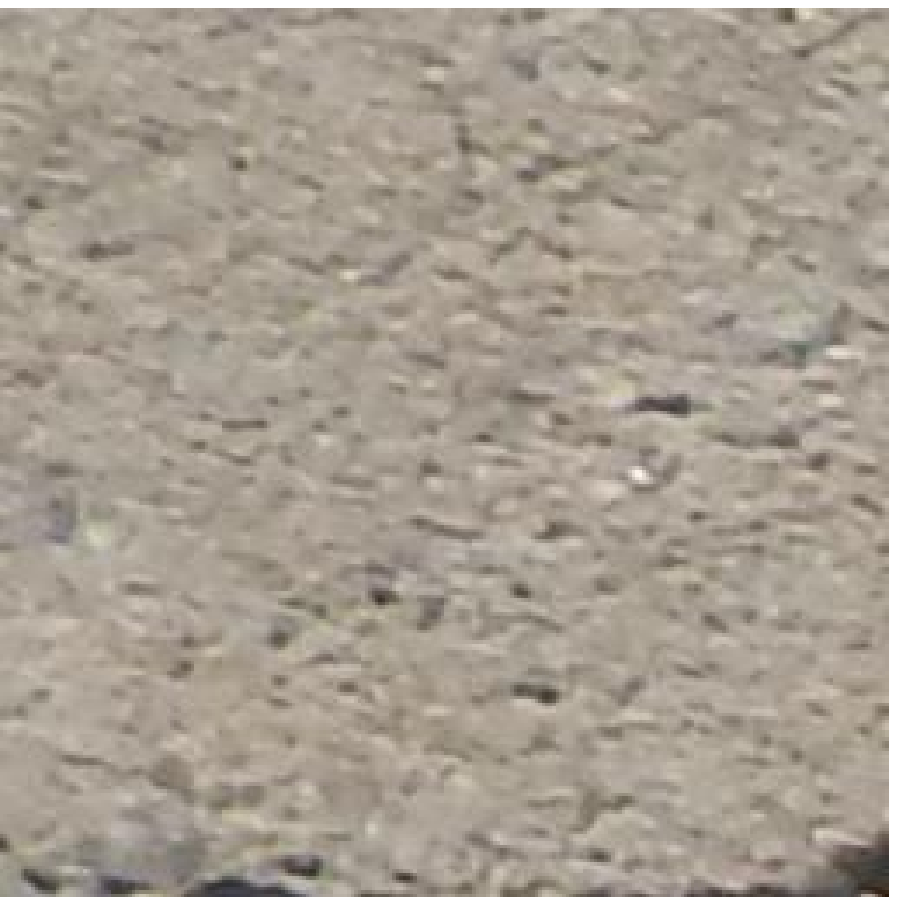}\label{fig:fig/error/CP/043-56.eps}}
\subfigure[Sample 5-2]{\includegraphics[scale=0.35]{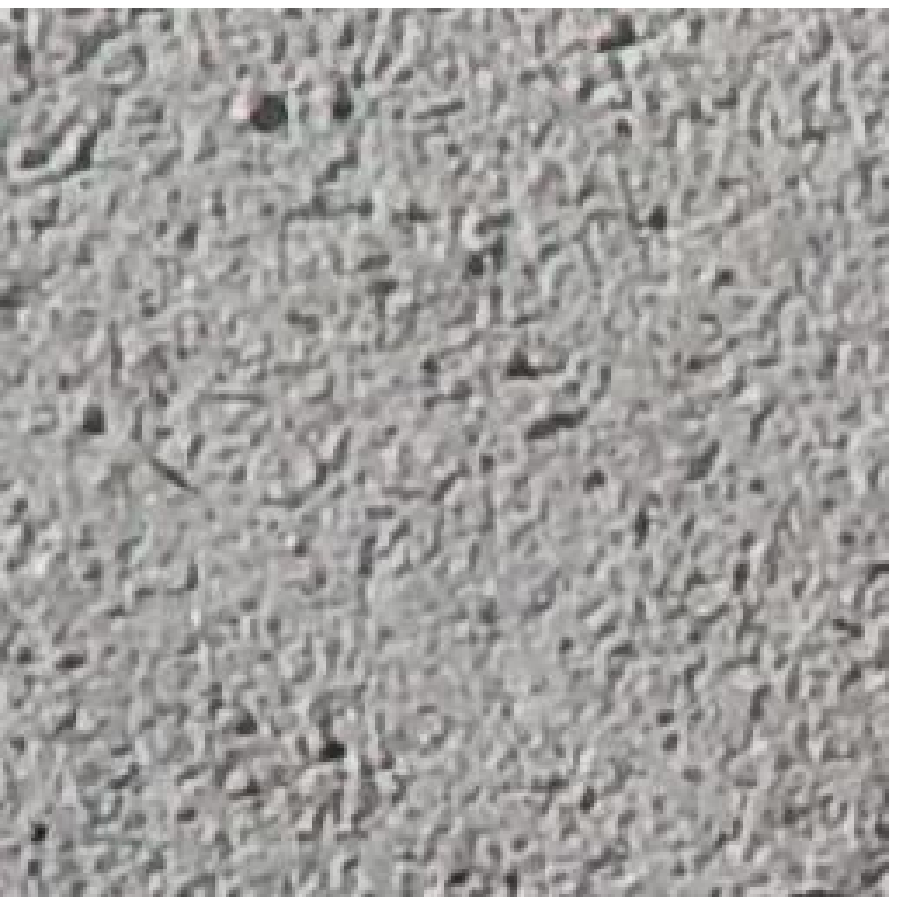}\label{fig:fig/error/CP/046-79.eps}}
\subfigure[Sample 5-3]{\includegraphics[scale=0.35]{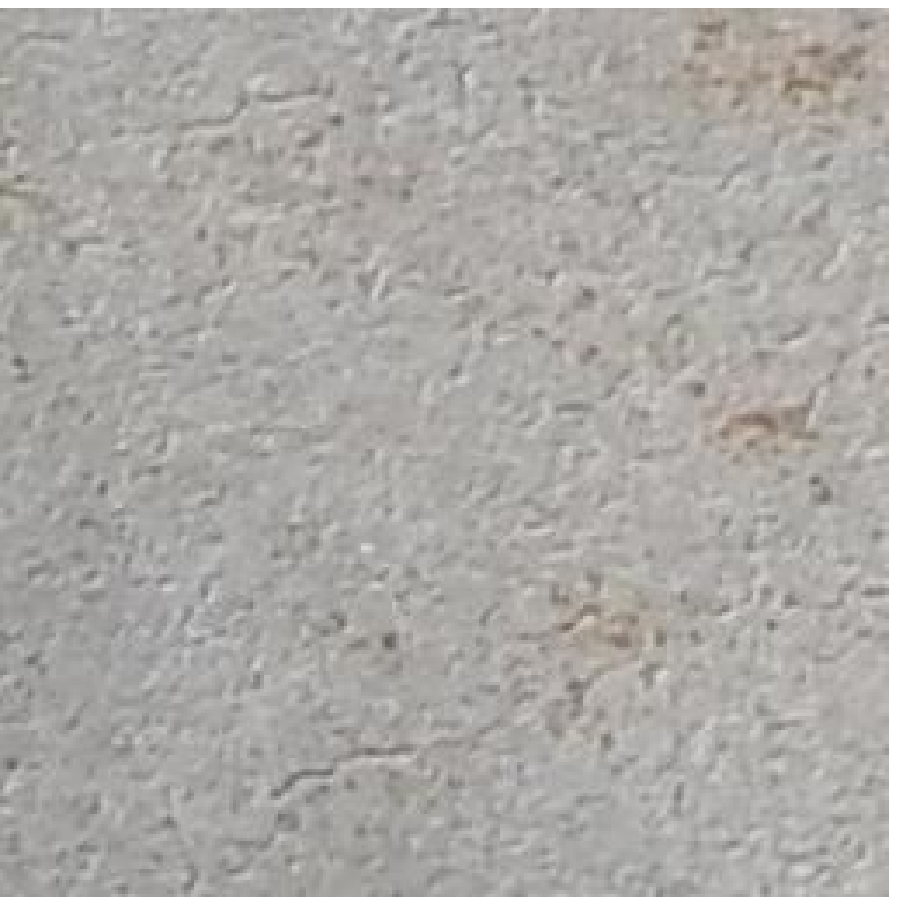}\label{fig:fig/error/CP/029-103.eps}}
\subfigure[Sample 5-4]{\includegraphics[scale=0.35]{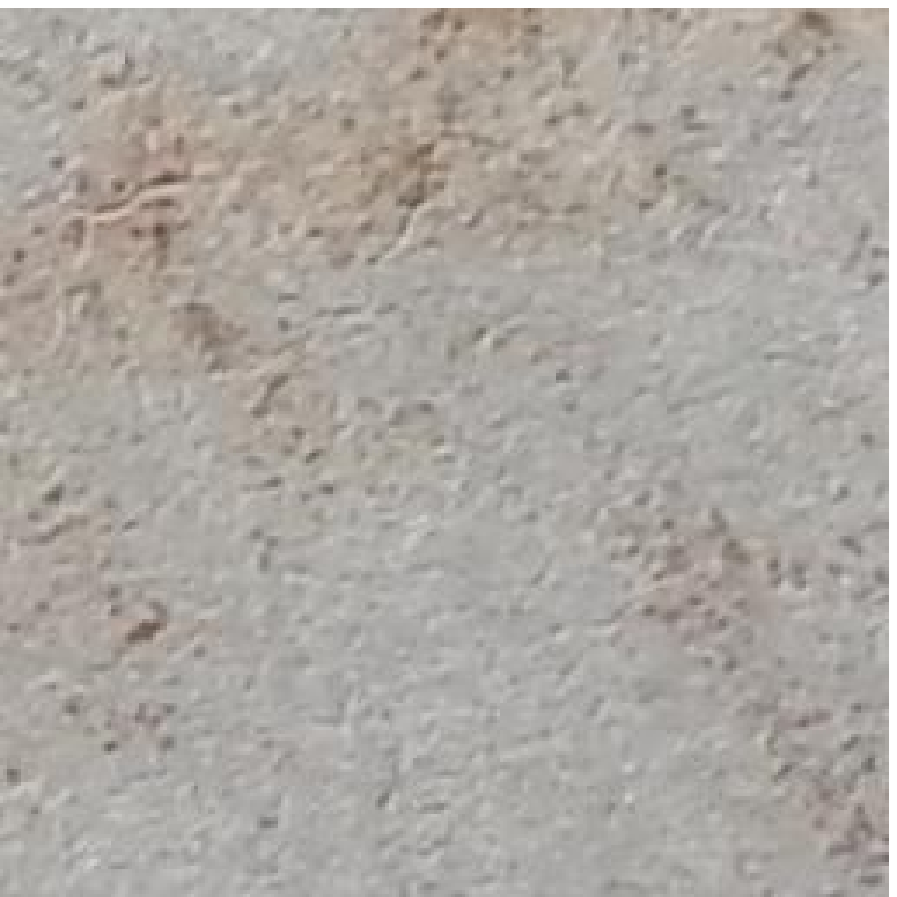}\label{fig:fig/error/CP/029-87.eps}}
\vspace{-2mm}
\caption{5) Pavement, Mis-judgment for cracked images}
\label{fig:case_CP_UP}
\vspace{-3mm}
\end{figure}

\begin{figure}[bt]
\centering
\subfigure[Sample 6-1]{\includegraphics[scale=0.35]{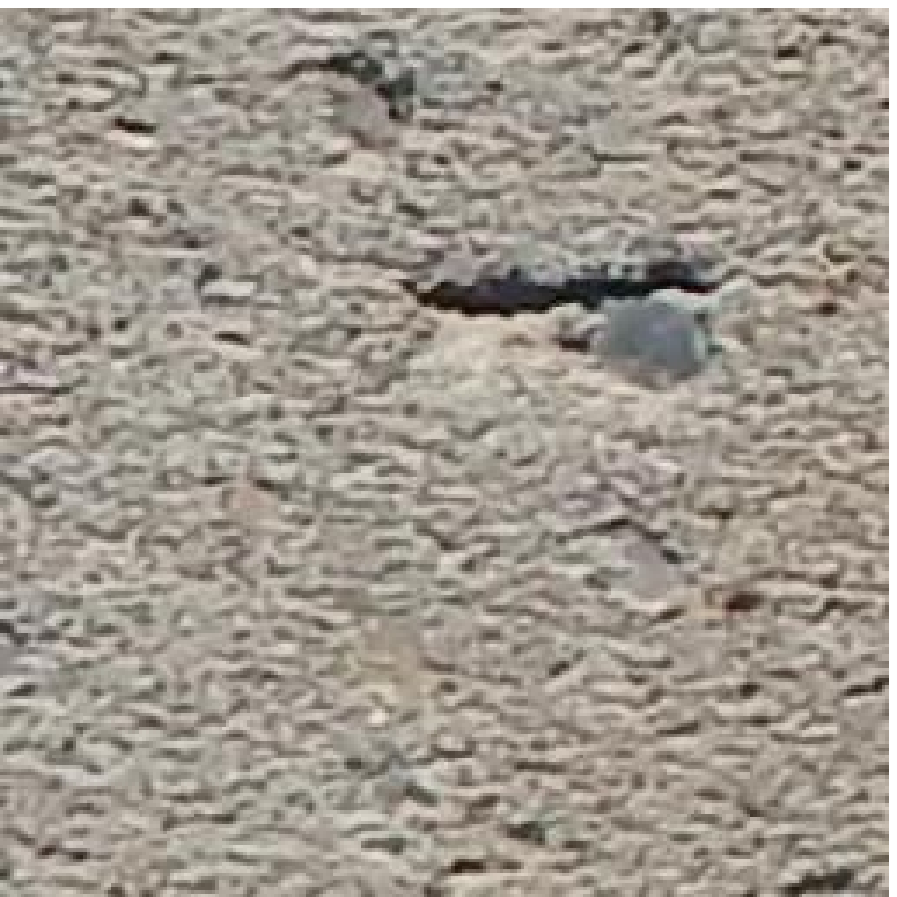}\label{fig:fig/error/UP/038-130.eps}}
\subfigure[Sample 6-2]{\includegraphics[scale=0.35]{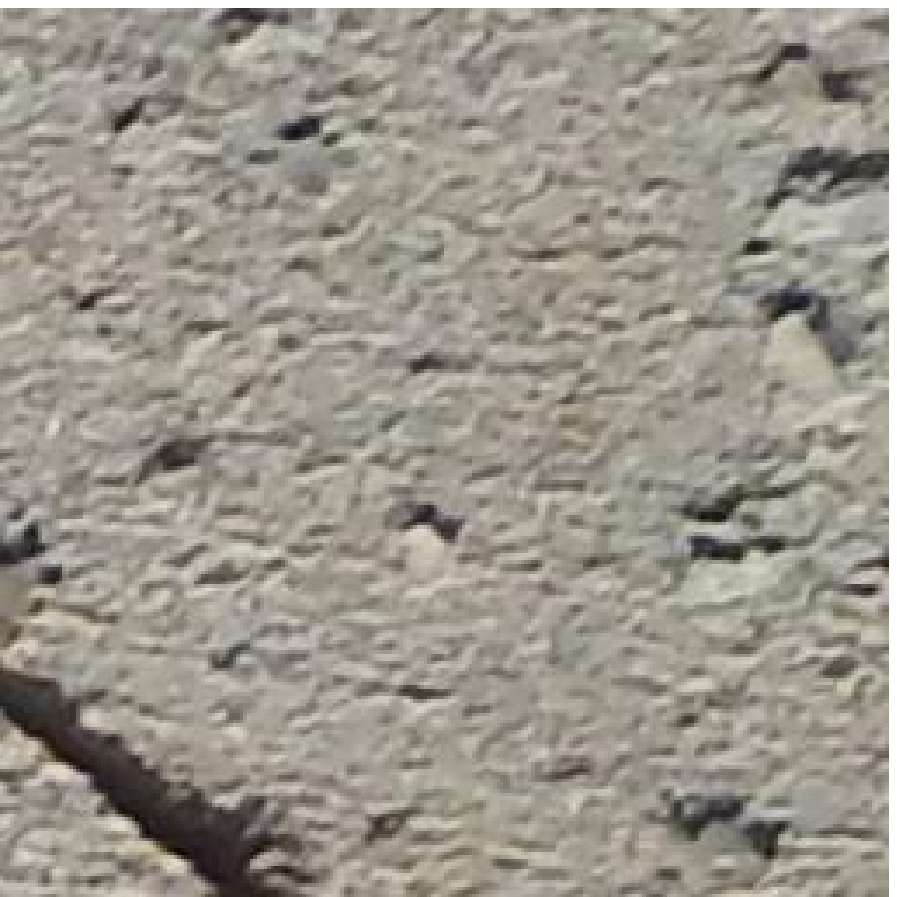}\label{fig:fig/error/UP/038-70.eps}}
\subfigure[Sample 6-3]{\includegraphics[scale=0.35]{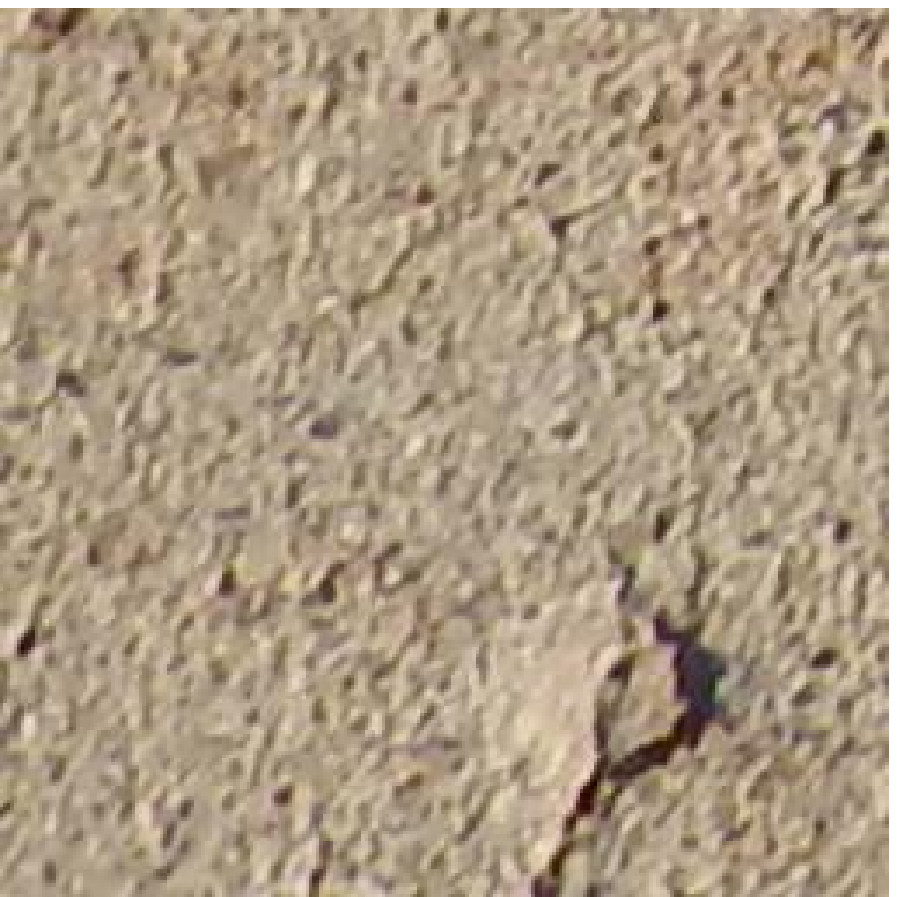}\label{fig:fig/error/UP/011-219.eps}}
\subfigure[Sample 6-4]{\includegraphics[scale=0.35]{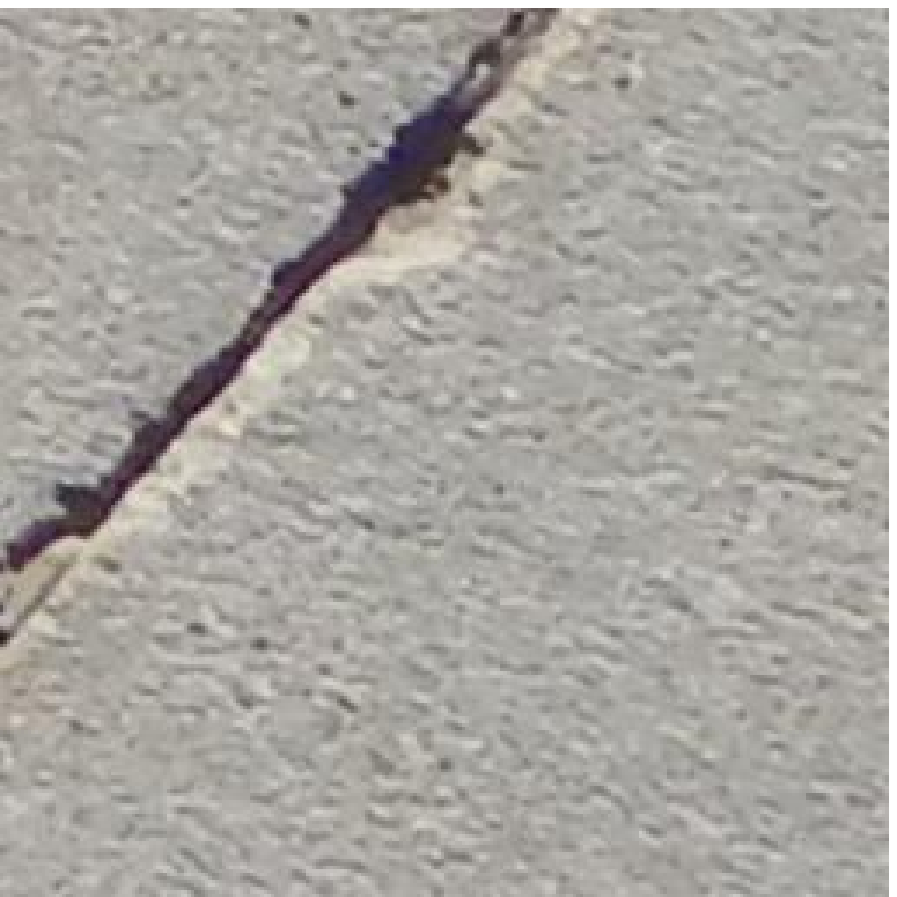}\label{fig:fig/error/UP/033-196.eps}}
\vspace{-2mm}
\caption{6) Pavement, Mis-judgment for images without cracks}
\label{fig:case_UP_CP}
\vspace{-3mm}
\end{figure}

\subsection{Mis-Classified data and Adversarial Examples}
\label{sec:AdversarialExamples}
As shown in Table~\ref{tab:classification_ratio1}, Adaptive DBN without fine-tuning method was not able to classify some cases about 4\% of test data images. The detailed investigation with respect to the cases showed that clearly judged cases of the presence or absence of cracks, such as the sample in Fig.~\ref{fig:data_sdnet_sample}, were correctly classified by Adaptive DBN. On the contrary, the cases that were not correctly classified were as follows: 1) image labeled as cracked, but data with few outstanding cracks and 2) image labeled as non-cracked, but data including features such as irregularities and small cracks. Fig.~\ref{fig:case_CD_UD} to Fig.~\ref{fig:case_UP_CP} show some of the images that were mis-classified in Table~\ref{tab:classification_ratio2}. We must consider that either these cases may be adversarial examples or not.

Adversarial examples are inputs to a neural network that leads a wrong answer. Some of these cases can be identified by humans but cannot be classified by deep learning systems. The labels are annotated by not only information from images but also knowledge from other perception. The expertise of civil engineers should be used to check the misclassification of the deep learning model.

The experts who can judge the concrete crack images investigated the 4\% mis-classified data by Adaptive DBN and they found half data of them were annotated as wrong label. The experts judged the data with the Japanese standard for concrete crack. Table~\ref{tab:concretecrackstandard} shows the Japanese concrete standard for cracks \cite{JapanConcreteStandard} and the experts use the standard. Adaptive DBN re-trains the modified data and then the classification accuracy increased for the mis-judgment data of test dataset.

The Adaptive DBN with the modified test dataset applied the training dataset, we found that such mis-labeled data are included in the training. SDNET2018 is an open benchmark dataset for concrete crack detection algorithm such as deep learning model. However, such adversarial examples may be included in the dataset and they will decrease classification capability. Unfortunately, no data detection algorithm to adversarial examples are without knowledge of experts. The method that the already trained network is used as the white box and the addition small perturb to the probability distribution to the original model is known as seen in ``Explaining and Harnessing adversarial examples \cite{Goodfellow2015}''. We will develop a novel method which automatically discovers adversarial examples and repair mis-annotated label during the training in future work.

\begin{table}[btp]
\caption{Concrete crack standard in Japanese \cite{JapanConcreteStandard}}
\vspace{-3mm}
\label{tab:concretecrackstandard}
\begin{center}
\scalebox{0.9}[0.9]{
\begin{tabular}{l|c|c|c|c|l}
\hline \hline
\multicolumn{2}{c|}{} & \multicolumn{3}{c|}{Durability} & \multirow{5}{13mm}{Waterproofing} \\ \cline{3-5}
\multicolumn{2}{l|}{Environment conditions} & & & &    \\ \cline{1-1}
\multicolumn{1}{l}{Degree of damage} & \diagbox[width=1.2cm,height=0.4cm]{}{}& Worse & Normal & Better & \\ \cline{1-1}
Category & & & & & \\ \hline \hline
\multirow{3}{23mm}{(A) Crack width that requires repair (mm)}& Large & 0.4 $\geq$ & 0.4 $\geq$ & 0.6 $\geq$ & 0.2 $\geq$ \\ 
& Middle & 0.4 $\geq$ & 0.6 $\geq$ & 0.8 $\geq$ & 0.2 $\geq$ \\
& Small & 0.6 $\geq$ & 0.8 $\geq$ & 1.0 $\geq$ & 0.2 $\geq$ \\ \hline
\multirow{3}{23mm}{(B) Crack width that doesn't require repair (mm)}& Large & 0.1 $\leq$ & 0.2 $\leq$ & 0.2 $\leq$  & 0.05 $\leq$ \\
& Middle & 0.1  $\leq$ & 0.2 $\leq$ & 0.3 $\leq$ & 0.05 $\leq$ \\
& Small & 0.2 $\leq$  & 0.3 $\leq$ & 0.3 $\leq$ & 0.05 $\leq$ \\ \hline
\hline
\end{tabular}
}
\end{center}
%\vspace{-5mm}
\end{table}

\section{Conclusive Discussion}
\label{sec:Discussion}
We focused on RBM and DBN, which are statistical models using the concept of likelihood, and proposed a structure-adaptive DBN that finds the optimal structure by generating / erasing neurons and hierarchization during learning. In this paper, the proposed model was applied to open data SDNET2018 on cracks in concrete structures. As a result of the experiment, it was possible to classify the test data of the three types of structures with accuracy of 94\% or more, which was higher than the CNN (the transfer learning with AlexNet). Among the cases that could not be classified, there are some cases that are mis-labeled about the presence or absence of cracks. In addition, we plan to develop a method that removes or repairs the adversarial cases in order to correctly classify cases that contain such mis-annotated cases.

\section*{Acknowledgment}
This work was supported by JSPS KAKENHI Grant Number 19K12142, 19K24365, and the obtained from the commissioned research by National Institute of Information and Communications Technology (NICT, 21405), JAPAN.

\end{document}